\documentclass[10pt,twocolumn,letterpaper]{article}

\usepackage{iccv}
\usepackage{times}
\usepackage{epsfig}
\usepackage{graphicx}
\usepackage{amsmath}
\usepackage{amssymb}
\usepackage{multirow}
\usepackage{pifont}
\usepackage{gensymb}
\usepackage{booktabs}
\usepackage{epsfig}
\usepackage{booktabs,multirow}
\usepackage{graphics}
\usepackage{threeparttable}
\usepackage{color}
\usepackage[normalem]{ulem}
\usepackage{multirow}
\usepackage{float}
\usepackage{amsfonts}
\usepackage{bm}
\usepackage{subfig}
\usepackage{enumitem}
\usepackage[numbers]{natbib}%sort&compress
\usepackage{array}
\usepackage[dvipsnames]{xcolor}
\usepackage{colortbl}
\usepackage[outercaption]{sidecap}
\usepackage{mathtools}
\usepackage{siunitx}
\usepackage[nopar]{lipsum}
\usepackage[export]{adjustbox}
\definecolor{myy}{RGB}{126,95,0}
\definecolor{mygray}{gray}{.9}
\definecolor{bblue}{RGB}{30,80,120}
\definecolor{mygray1}{gray}{.7}
\usepackage{bm}
\usepackage{mathtools}
\usepackage{siunitx}
\usepackage[nopar]{lipsum}

\newcolumntype{I}{!{\vrule width 1pt}}

\definecolor{ggray}{RGB}{127,127,127}

\makeatletter
\newcommand{\thickhline}{%
   \noalign {\ifnum 0=`}\fi \hrule height 1pt
   \futurelet \reserved@a \@xhline
}
\makeatother

\usepackage{caption}
\usepackage{algorithm}
\usepackage{listings}

\usepackage{etoolbox}
\makeatletter
\AfterEndEnvironment{algorithm}{\let\@algcomment\relax}
\AtEndEnvironment{algorithm}{\kern2pt\hrule\relax\vskip3pt\@algcomment}
\let\@algcomment\relax
\newcommand\algcomment[1]{\def\@algcomment{\footnotesize#1}}
\renewcommand\fs@ruled{\def\@fs@cfont{\bfseries}\let\@fs@capt\floatc@ruled
  \def\@fs@pre{\hrule height.8pt depth0pt \kern2pt}%
  \def\@fs@post{}%
  \def\@fs@mid{\kern2pt\hrule\kern2pt}%
  \let\@fs@iftopcapt\iftrue}
\makeatother

\usepackage{mathtools, nccmath}
\usepackage{comment}
\usepackage{xspace}
\usepackage{ragged2e}
\newcolumntype{P}[1]{>{\RaggedRight\hspace{0pt}}p{#1}}
\newcolumntype{X}[1]{>{\RaggedRight\hspace*{0pt}}p{#1}}
\usepackage{tcolorbox}
\usepackage[edges]{forest}
\colorlet{linecol}{black!75}
\usepackage{xkcdcolors} % xkcd colors
\colorlet{mhpurple}{Plum!80}
\definecolor{mygray2}{gray}{.6}
\definecolor{mygray3}{gray}{.3}
\definecolor{mygray}{gray}{.9}
\definecolor{mywarning}{RGB}{233,144,61}
\definecolor{ggray}{RGB}{127,127,127}
\definecolor{reda}{RGB}{192,0,0}
\definecolor{redb}{RGB}{217,148,143}
\definecolor{myyellow}{RGB}{190,144,0}
\definecolor{mygreen}{RGB}{0,153,0}
\definecolor{mygreen2}{RGB}{153,255,153}
\definecolor{myred}{RGB}{177,35,15}
\definecolor{myblue}{RGB}{58,79,116}
\definecolor{myy}{RGB}{254,203,50}
\definecolor{myy2}{RGB}{191,144,0}
\definecolor{myg}{RGB}{205,205,205}
\definecolor{myg2}{RGB}{80,80,80}
\definecolor{codegreen}{RGB}{79,126,127}
\definecolor{codedefine}{RGB}{153,54,159}
\definecolor{codefunc}{RGB}{73,122,234}
\definecolor{codecall}{RGB}{73,122,234}
\definecolor{codepro}{RGB}{212,96,80}
\definecolor{codedim}{RGB}{89,152,195}

\usepackage[pagebackref=true,breaklinks=true,colorlinks,bookmarks=false]{hyperref}
\usepackage[utf8]{inputenc}

\usepackage{cleveref}
\crefname{section}{§}{§§}
\Crefname{section}{§}{§§}

\iccvfinalcopy 

\def\iccvPaperID{4528} % *** Enter the ICCV Paper ID here

\ificcvfinal\fi

\begin{document}

%%%%%%%%% TITLE
\title{Large-Scale Person Detection and Localization using Overhead Fisheye Cameras}

\author{
Lu Yang\textsuperscript{1}\footnotemark[1] , Liulei Li\textsuperscript{2}\footnotemark[1] , Xueshi Xin\textsuperscript{1} , Yifan Sun\textsuperscript{3}~, Qing Song\textsuperscript{1} , Wenguan Wang\textsuperscript{4}\footnotemark[2]\\
\small \textsuperscript{1} Beijing University of Posts and Telecommunications~~\textsuperscript{2} ReLER, AAII, University of Technology Sydney\\
\small \textsuperscript{3} Baidu~~\textsuperscript{4} ReLER, CCAI, Zhejiang University\\
\small\url{https://LOAFisheye.github.io/}
}

\maketitle
\ificcvfinal\thispagestyle{empty}\fi

%%%%%%%%% ABSTRACT
\begin{abstract}
   %partitioning Harnessing
%Identifying the correct location of persons
% Location determination  is a cornerstone of many real-world applications.  finds wide applications in daily life.
\footnotetext[1]{The first two authors contribute equally to this work.}\footnotetext[2]{Corresponding author: Wenguan Wang.}
Location determination finds wide applications in daily life. Instead of existing efforts devoted to localizing tourist photos$_{\!}$ captured$_{\!}$ by$_{\!}$ perspective$_{\!}$ cameras, in this article,$_{\!}$~we focus$_{\!}$ on$_{\!}$ devising$_{\!}$ person$_{\!}$ positioning$_{\!}$ solutions$_{\!}$ using$_{\!}$ over- head fisheye cameras. Such solutions are advantageous in large field of view (FOV), low cost, anti-occlusion, and~un- aggressive$_{\!}$ work$_{\!}$ mode$_{\!}$ (without$_{\!}$ the$_{\!}$ necessity$_{\!}$ of$_{\!}$ cameras$_{\!}$ car-$_{\!}$ ried by$_{\!}$ persons). However, related studies are quite scarce, due to the paucity of data. To stimulate research in this exciting area, we present LOAF, the first \underline{l}arge-scale \underline{o}verhe\underline{a}d \underline{f}isheye dataset for person detection and localization. LOAF is built with many essential features, \eg, i) the data cover abundant diversities in scenes, human pose, density, and location; ii) it contains currently the largest number of annotated pedestrian, \ie, 457K bounding boxes with ground-truth location information; iii) the body-boxes are labeled as radius-aligned so as to fully address the positioning challenge. To approach localization, we build a fisheye person detection network, which exploits the fisheye distortions by a rotation-equivariant training  strategy and predict radius-aligned human boxes end-to-end. Then, the actual locations of the detected persons are calculated by a numerical solution on the fisheye model and camera altitude data. Extensive experiments on LOAF validate the superiority of our fisheye detector \wrt previous methods, and show that our whole fisheye positioning solution is able to locate all persons in FOV with an accuracy of 0.5~m, within 0.1~s.
\end{abstract}

   \vspace{-10pt}
%%%%%%%%% BODY TEXT
\section{Introduction}
Accurate position finding of persons attracts growing interest from both research and industrial
communities, since it$_{\!}$ plays$_{\!}$ a$_{\!}$ crucial$_{\!}$ role$_{\!}$ in$_{\!}$ numerous$_{\!}$ location-sensitive$_{\!}$ applica- tion$_{\!}$ scenarios$_{\!}$ (\eg,$_{\!}$ surveillance,$_{\!}$ smart$_{\!}$  home,$_{\!}$ public$_{\!}$ health). 
Nevertheless, due to the line-of-sight (LOS) issue, GPS is unreliable$_{\!}$ in$_{\!}$ interior$_{\!}$ spaces$_{\!}$ and$_{\!}$ urban$_{\!}$ canyon.$_{\!}$ 
To$_{\!}$ overcome such$_{\!}$ limitation,$_{\!}$ various$_{\!}$ alternative$_{\!}$ solutions$_{\!}$ are$_{\!}$ investigated.$_{\!}$ \textit{Signal$_{\!}$ based}$_{\!}$ solutions,$_{\!}$ including$_{\!}$ Bluetooth$_{\!}$~\cite{chen2013bayesian}$_{\!}$ and$_{\!}$  Wi-Fi~\cite{yang2015wifi}, are popular, but they are easily interfered by changing environments and nearby human bodies~\cite{xiao2018indoor}. A complementary stream of work is \textit{vision based}; they typically make use of traditional cameras, RGBD cameras, or in-built smartphone cameras, and enjoy the advantage of reliable services.  To get location information, visual positioning solutions usually refer to a pre-acquired 3D map or a  geo-tagged database as the scene representation~\cite{taira2018inloc}, or directly utilize the captured image to estimate the camera pose~\cite{kendall2015posenet}.

\begin{figure}
   \begin{center}
   \vspace{-10pt}
      \includegraphics[width=\linewidth]{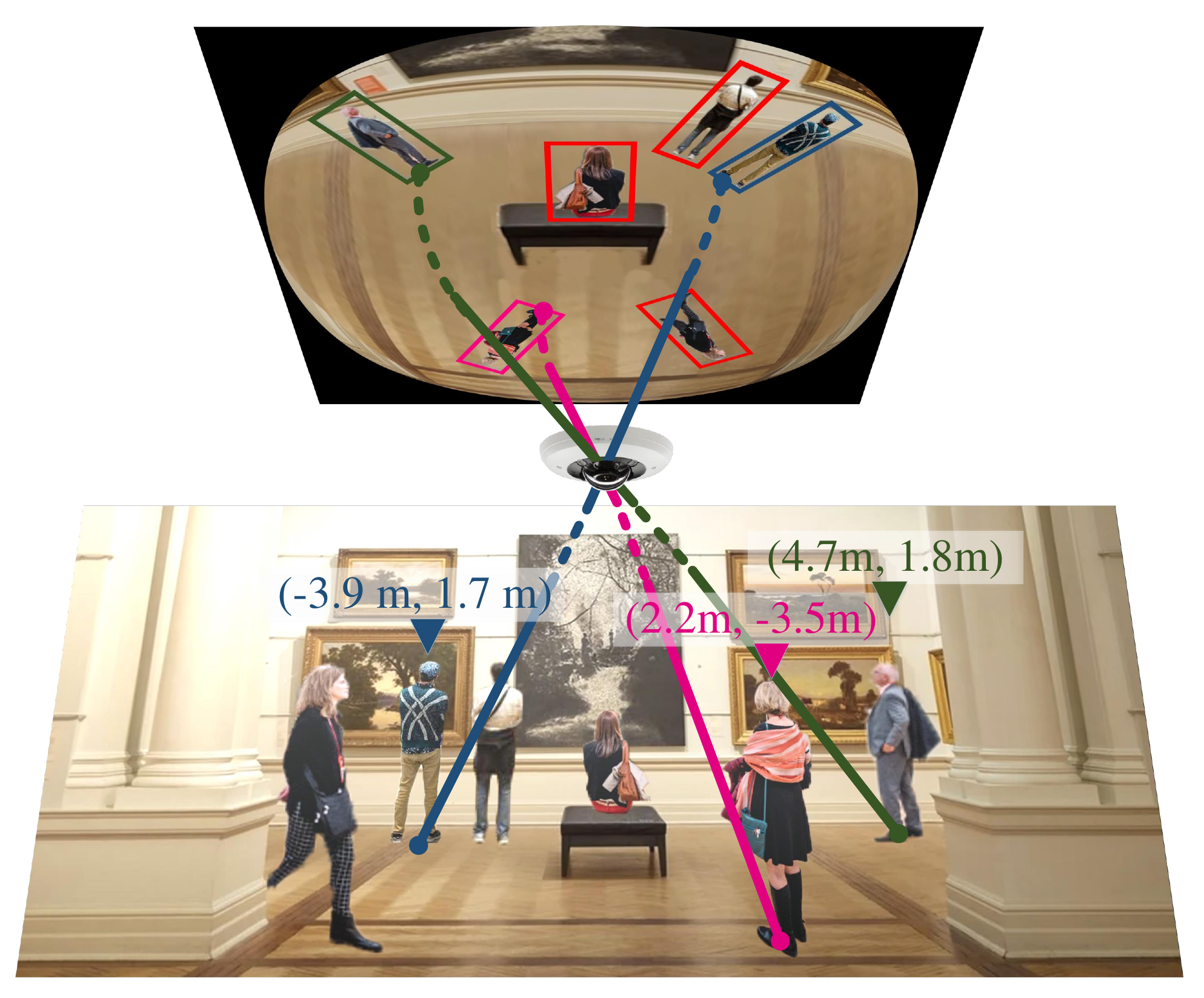}
   \end{center}
   \vspace{-15pt}
   \captionsetup{font=small}
   \caption{{\small{\textbf{$_{\!}$Person$_{\!}$ positioning$_{\!}$ using$_{\!}$ overhead$_{\!}$ fisheye$_{\!}$ camera}:$_{\!}$~We detect humans on omnidirectional images and then project detects onto the real world coordinates to obtain physical locations. Compared with using perspective cameras, our fisheye camera based solution is favored in low cost, high accuracy, and fast speed.}}}
   \label{fig:fig1}
   \vspace{-15pt}
\end{figure}

Although visual localization has been a hotspot issue for many years, existing efforts are mainly dedicated to urban place recognition or indoor camera localization, based on \textit{perspective cameras}~\cite{taira2018inloc,kendall2015posenet,sun2017dataset,li2012worldwide}. None of them addresses {person positioning} by using \textit{overhead fisheye cameras}, even though fisheye cameras are widely used in visual surveillance applications. One possible reason is the lack of accessible datasets, compounded by considerable costs involved in data collection. In this article, we provide a \underline{l}arge-scale \underline{o}verhe\underline{a}d \underline{f}isheye dataset, LOAF, for person detection and localization in both indoor and outdoor scenes. 

$_{\!}$Compared$_{\!}$ with$_{\!}$ perspective$_{\!}$ cameras, overhead$_{\!}$ (top-view)$_{\!}$ fisheye$_{\!}$ cameras$_{\!}$ are promoted due to less occlusion among people$_{\!}$ and$_{\!}$ larger$_{\!}$ field$_{\!}$ of$_{\!}$ view$_{\!}$ (FOV) -- allowing$_{\!}$ the$_{\!}$ coverage$_{\!}$ of$_{\!}$ a$_{\!}$ large$_{\!}$ space$_{\!}$ using$_{\!}$ a$_{\!}$ single, low-cost$_{\!}$ camera. Only$_{\!}$ few public datasets~\cite{del2021robust,li2019supervised,duan2020rapid} provide top-view fisheye data. Unfortunately, they only cover quite few scenes and their people-box annotations do not adequately address the positioning challenge, negatively affecting location approximation (see \S\ref{sec:dataset_annotation} for detailed analysis). Differently, LOAF specifically targets at person localization in surveillance applications, and has the following appealing characteristics: %iscriminative feature
\begin{itemize}[leftmargin=*]
   \setlength{\itemsep}{0pt}
   \setlength{\parsep}{-2pt}
   \setlength{\parskip}{-0pt}
   \setlength{\leftmargin}{-8pt}
   \vspace{-6pt}
   \item \textit{Large-scale}: LOAF is the largest in the filed, to our best knowledge. It consists of over 70 videos, with more than 43K frame images, 457K person-detection annotations as well as corresponding location information.

   \item  \textit{High diversity}: LOAF contains a wide variety of surveillance scenarios; it$_{\!}$ includes$_{\!}$ a$_{\!}$ total$_{\!}$ of$_{\!}$ 11$_{\!}$ indoor$_{\!}$ and$_{\!}$ 40$_{\!}$ out- door scenes, and the data are captured at different times of day and cover different illumination conditions.

   \item \textit{Positioning-aware person-box annotation}: LOAF offers radius-aligned human-box annotations. Compared$_{\!}$ with$_{\!}$ other$_{\!}$ person$_{\!}$ representations, \ie, head$_{\!}$ center$_{\!}$~\cite{del2021robust}, axis- aligned$_{\!}$~\cite{seidel2019improved} or$_{\!}$ body-aligned$_{\!}$ box$_{\!}$~\cite{li2019supervised,duan2020rapid}, used in previous fisheye datasets, the radius-aligned box is promoted, as it fits well radially-oriented bodies~\cite{li2019supervised}, and, more importantly, it is better aware of the positioning problem.$_{\!\!\!}$
   \vspace{-4pt}
\end{itemize}

$_{\!}$Moreover,$_{\!}$ we$_{\!}$ devise$_{\!}$ a$_{\!}$ person$_{\!}$ positioning$_{\!}$ system$_{\!}$ that$_{\!}$ first detects persons from raw fisheye images, and then calcula- tes physical locations based on fisheye visual model and al- titude$_{\!}$ information$_{\!}$ (see$_{\!}$ Fig.\!~\ref{fig:fig1}).$_{\!}$ Clearly,$_{\!}$ high-quality$_{\!}$ person$_{\!}$ detection is the crucial premise for precise localization using overhead fisheye cameras. Fisheye lenses provide large FOV, at the cost of strong radial distortion.  This~makes~pe- destrian detection in top-view fisheye images a much harder task, compared with using perspective cameras. Studies on overhead fisheye pedestrian detection are very scarce~\cite{seidel2019improved,li2019supervised,rashed2021generalized,duan2020rapid,tamura2019omnidirectional}; they rarely concern the link with person positioning, and many of them~\cite{seidel2019improved,li2019supervised} are even not trainable due to the lack of fisheye data. With our LOAF dataset,~we develop$_{\!}$ a$_{\!}$ novel$_{\!}$ query$_{\!}$ based$_{\!}$ fisheye$_{\!}$ human$_{\!}$ detector.$_{\!}$ It$_{\!}$ expli- citly exploits fisheye geometry by accommodating \textit{rotation equivariance} into the matching between queries and human instances during network training. The insight here~is~intui- tive:$_{\!}$  for$_{\!}$  a$_{\!}$  robust$_{\!}$ fisheye$_{\!}$ detector,$_{\!}$ rotation$_{\!}$ of$_{\!}$ a$_{\!}$ fisheye$_{\!}$  image should result in correspondingly rotated detections. In addition,$_{\!}$ our$_{\!}$ detection$_{\!}$ algorithm$_{\!}$ learns$_{\!}$ to$_{\!}$ predict radius-aligned person boxes,  facilitating localization estimation.

We test our fisheye person detection algorithm as well as our whole positioning system over LOAF. We find that our detector significantly outperforms previous methods, and our full system delivers precise localization results. We also empirically show that our algorithm generalizes well on previous top-view fisheye person detection datasets~\cite{duan2020rapid,tezcan2022wepdtof}.

\vspace{-4pt}
\section{Related Work}
\vspace{-4pt}

\noindent\textbf{Accurate$_{\!}$ Positioning.}$_{\!}$ GPS$_{\!}$ is$_{\!}$ the$_{\!}$ most$_{\!}$ popular$_{\!}$ system$_{\!}$ for outdoor localization. As it requires LOS between the satellites and the handset, GPS$_{\!}$ does$_{\!}$ not$_{\!}$ function well in urban canyons, indoors$_{\!}$ and$_{\!}$ basements\!~\cite{gu2009survey,farid2013recent}.$_{\!}$
This$_{\!}$ triggered$_{\!}$ the development of alternative positioning solutions, following a$_{\!}$ multi-disciplinary$_{\!}$ approach.$_{\!}$ Concretely,$_{\!}$ there$_{\!}$ are$_{\!}$ two$_{\!}$ main$_{\!}$ schemes of the alternatives: \textit{signal-based} and \textit{vision-based}.

Signal-based$_{\!}$ positioning$_{\!}$ systems$_{\!}$ typically$_{\!}$~lean$_{\!}$ on sound wave$_{\!}$~\cite{qi2017robust}, geomagnetism$_{\!}$~\cite{haverinen2009global}, radio$_{\!}$ frequency$_{\!}$ (RF)$_{\!}$~\cite{farid2013recent,yang2015wifi,saab2010standalone,chen2013bayesian,alarifi2016ultra}, and infrared radiation (IR)$_{\!}$~\cite{want1992active}, as well as different location determination techniques, such as TOA (time~of arrival)$_{\!}$~\cite{gu2009survey} and RSS (received signal strength)$_{\!}$~\cite{farid2013recent}. 
The main challenge to signal-based systems is the sensitivity~to environment$_{\!}$ changes,$_{\!}$ such$_{\!}$ as$_{\!}$ object$_{\!}$ moving,$_{\!}$ diffraction$_{\!}$ and reflection, which affect signal propagation~\cite{farid2013recent}. 

Visual$_{\!}$ data$_{\!}$ is$_{\!}$ another$_{\!}$ potential$_{\!}$ information$_{\!}$ source$_{\!}$ for$_{\!}$ pre- cise$_{\!}$ localization.$_{\!}$  Since$_{\!}$  put$_{\!}$ forward$_{\!}$~\cite{shafer1998new},$_{\!}$ visual$_{\!}$ positioning$_{\!}$ has$_{\!}$ became$_{\!}$ a$_{\!}$ hot$_{\!}$ topic$_{\!}$ in$_{\!}$ robotics$_{\!}$ and$_{\!}$ computer$_{\!}$ vision.$_{\!}$~Some methods$_{\!}$ adopt$_{\!}$ a$_{\!}$ pre-built$_{\!}$ geo-tagged$_{\!}$ database$_{\!}$ or$_{\!}$ a$_{\!}$ 3D$_{\!}$ scene
 model,$_{\!}$ as$_{\!}$ the$_{\!}$ reference$_{\!}$ for$_{\!}$ camera$_{\!}$ location$_{\!}$ estimation$_{\!}$~\cite{hays2008im2gps,zamir2010accurate,vaca2012city,zamir2014image,yousif2017monorgbd}.$_{\!}$
Some$_{\!}$ others$_{\!}$ rely$_{\!}$ on$_{\!}$ recognizing$_{\!}$ some$_{\!}$~dep- loyed$_{\!}$ coded$_{\!}$ targets$_{\!}$~\cite{fiala2005artag,schweighofer2006robust,degol2017chromatag,hu2019deep},$_{\!}$ \eg,$_{\!}$ concentric rings, barcodes,$_{\!}$ colored$_{\!}$ dots,$_{\!}$ \etc.$_{\!}$ Some$_{\!}$ recent$_{\!}$ ones$_{\!}$ utilize$_{\!}$ deep learning techniques to replace some components (\eg, image retrieval, descriptor matching) in traditional systems\!~\cite{boniardi2019robot,ch2020resolving}, or regress the camera pose directly\!~\cite{kendall2015posenet,kendall2017geometric,bresson2019urban}. Though promising, existing visual positioning systems are mostly founded on perspective cameras.

Due to the problem complexity, there is no persuasive solution for precise positioning yet, and hybrid schemes are often applied in practice. Fisheye cameras have advantage of providing wide FOV with low cost and reduced occlusion, while the research landscape is sparse for fisheye camera based localization~\cite{zhu2019object,del2021robust}. To foster research in this direction and facilitate practical deployment, we contribute the first overhead fisheye dataset, to our knowledge, that allows to conduct the task of person localization at large scale.

\noindent\textbf{Deep Learning based Visual (Camera) Localization.}~Re- garding$_{\!}$ application$_{\!}$ scenarios,$_{\!}$ scholars$_{\!}$ in$_{\!}$ computer$_{\!}$~vision community are mainly aware of \textit{city-scale location recogni- tion}~\cite{arandjelovic2016netvlad,weyand2016planet,sattler2017large,sattler2018benchmarking} and \textit{indoor camera localization}~\cite{brachmann2017dsac,walch2017image}. 
Popular visual localization approaches can be divided$_{\!}$ as$_{\!}$ %into three groups:
\textit{retrieval-based},$_{\!}$ \textit{regression-based},$_{\!}$ and \textit{structure-based}, according to the camera pose estimation strategy. Retrieval-based methods~\cite{sattler2017large,balntas2018relocnet,ding2019camnet,yang2019sanet,sarlin2021back} represent a scene as a database of geo-tagged photos and use geo-tag of the most relevant database photo as an approximation to the camera position~\cite{sattler2017large}.  Regression-based methods learn to encode the scene into a deep network and directly regress a 6DOF camera pose~\cite{kendall2015posenet,walch2017image,kendall2017geometric} from a captured image. Structure-based methods$_{\!}$~\cite{brachmann2019expert,sarlin2019coarse,germain2021neural} pose the localization$_{\!}$ problem$_{\!}$ as$_{\!}$ a$_{\!}$ camera$_{\!}$ resectioning$_{\!}$ task$_{\!}$~\cite{sattler2017large}.$_{\!}$ They$_{\!}$ first represent scenes via 3D models and establish a set of 2D-3D matches, then recover the full camera pose by employing a PnP solver~\cite{hesch2011direct} inside a RANSAC~\cite{fischler1981random} loop.

These visual positioning techniques seek to localize images captured by handheld devices~\cite{li2012worldwide,chen2011city} or vehicle cameras~\cite{maddern20171} and demand pre-created geo-tagged databases or 3D maps. And they work on an \textit{active} mode -- users need to carry the cameras. Differently, we address person localization in surveillance scenarios, by using {stationary}, overhead fisheye cameras. This yields a \textit{passive}, low-cost scheme, where the localization is completed by a numerical solution on the fisheye model fused with altitude information. Thus our scheme faces a different challenge, \ie, conducting robust person detection from non-rectilinear fisheye images. Overall, our scheme fills the gap left by conventional studies and is complementary to existing positioning systems.

\noindent\textbf{Person$_{\!}$ Detection$_{\!}$ in$_{\!}$ Overhead$_{\!}$ Fisheye$_{\!}$ Images.} As$_{\!}$ a$_{\!}$ ca- nonical$_{\!}$ subproblem of object detection, pedestrian detection has long received great interest owing to its broad applications such as intelligent surveillance and autonomous vehicles\!~\cite{chen2021deep,cao2021handcrafted}. By contrast, people detection in overhead fisheye images has been studied much less, due to the absence of such datasets. Moreover, adapting standard pedestrian detectors to top-view fisheye cameras is difficult~\cite{chiang2021efficient}: First, the appearance of people in fisheye images is arbitrary-oriented, while in perspective images, people typically appear upright. Second, people suffer from severe geometric distortions, particularly in the fisheye image's periphery.

Faced with the above challenges, early attempts make use of handcrafted features (\eg, HOG, LBP) as well as standard pedestrian detectors with slight modifications to account$_{\!}$ for$_{\!}$ fisheye$_{\!}$ geometry~\cite{krams2017people,chiang2014human,cinaroglu2014direct,wang2017template}. The$_{\!}$ common$_{\!}$ approach is dewarping fisheye images\!~\cite{chiang2014human} or features\!~\cite{krams2017people} so as to approximate normal people' appearances from the deformed ones. Though this simplifies the classification of features, approximation errors  are inevitable, causing performance degradation\!~\cite{tamura2019omnidirectional}. Later, a few CNN-based fisheye detectors were developed. Some of them are \textit{training-free}. For example, \cite{seidel2019improved} runs standard YOLOv2\!~\cite{redmon2017yolo9000} on dewarped versions of overlapped windows extracted from a fisheye image.$_{\!}$ Li \etal\!~\cite{li2019supervised}$_{\!}$ rotate$_{\!}$ a$_{\!}$ fisheye$_{\!}$ image$_{\!}$ in$_{\!}$ 15$^\circ$$_{\!}$ increments, and apply off-the-shelf YOLOv3\!~\cite{redmon2018yolov3} only to the top-center region of each rotated image, where people usually appear upright. Some more recent methods are \textit{trainable}: \cite{tamura2019omnidirectional} trains YOLOv2 with rotated perspective images so as to handle omnidirectional images without test-time transformation; \cite{duan2020rapid} trains YOLOv3 with human-aligned boxes.

Although many prior arts$_{\!}$~\cite{seidel2019improved,li2019supervised,tamura2019omnidirectional} allowing to detect persons$_{\!}$ without$_{\!}$ any$_{\!}$ fisheye$_{\!}$ training$_{\!}$ data,$_{\!}$ they$_{\!}$ require$_{\!}$ a$_{\!}$ cer- tain$_{\!}$ amount$_{\!}$ of$_{\!}$ computation$_{\!}$ time$_{\!}$ for post-processing:$_{\!}$ \cite{seidel2019improved}$_{\!}$ carries$_{\!}$ out$_{\!}$ detection$_{\!}$ in$_{\!}$ multiple$_{\!}$ perspective$_{\!}$ images$_{\!}$ dewarped from one omnidirectional image; % and the detection performance is affected by transformation accuracy
\cite{li2019supervised}$_{\!}$ applies$_{\!}$ YOLOv3$_{\!}$ 24$_{\!}$ times$_{\!}$ to$_{\!}$ each$_{\!}$ fisheye$_{\!}$ image;$_{\!}$
\cite{tamura2019omnidirectional}$_{\!}$ needs$_{\!}$ a$_{\!}$ grouping$_{\!}$ process$_{\!}$ to$_{\!}$ eliminate$_{\!}$ numerous$_{\!}$ redundant$_{\!}$ %and$_{\!}$ false$_{\!}$ positive$_{\!}$
results,$_{\!}$ caused$_{\!}$ by$_{\!}$ a$_{\!}$ rotation-invariant$_{\!}$ training$_{\!}$ strategy.$_{\!}$ Hence,$_{\!}$ the$_{\!}$ utility$_{\!}$ of$_{\!}$ previous$_{\!}$ me- thods$_{\!}$  is$_{\!}$ limited$_{\!}$ in$_{\!}$ the$_{\!}$ context$_{\!}$ of$_{\!}$ visual$_{\!}$ positioning.$_{\!}$ Our$_{\!}$ ana- lysis$_{\!}$ in$_{\!}$ \S\ref{sec:dataset_annotation}$_{\!}$ sheds$_{\!}$ light$_{\!}$ on$_{\!}$ the$_{\!}$ weaknesses$_{\!}$ of$_{\!}$ human representations adopted by existing$_{\!}$ fisheye$_{\!}$ detectors$_{\!}$ (\eg,  head$_{\!}$ center$_{\!}$~\cite{del2021robust}, horizontal$_{\!}$ or$_{\!}$ body-aligned$_{\!}$ box$_{\!}$~\cite{seidel2019improved,li2019supervised,duan2020rapid})$_{\!}$~in regard$_{\!}$ to$_{\!}$ localization.$_{\!}$ Thus$_{\!}$ we$_{\!}$ supply$_{\!}$ our$_{\!}$ large-scale$_{\!}$ dataset with$_{\!}$ positioning-aware$_{\!}$ person-detection$_{\!}$ annotation.$_{\!}$ As$_{\!}$ a$_{\!}$~re-
sult,$_{\!}$ our$_{\!}$ human$_{\!}$ detector$_{\!}$ can$_{\!}$ benefit$_{\!}$ from$_{\!}$ end-to-end,$_{\!}$ fisheye$_{\!}$ visual$_{\!}$ pattern$_{\!}$ learning$_{\!}$ and$_{\!}$ output$_{\!}$ radius-aligned$_{\!}$ body-boxes$_{\!}$ for$_{\!}$ precise$_{\!}$ positioning.$_{\!}$ Further,$_{\!}$ by$_{\!}$ regularizing$_{\!}$ the$_{\!}$ learning$_{\!}$ of$_{\!}$ image$_{\!}$ representations$_{\!}$ and$_{\!}$ instance$_{\!}$ queries$_{\!}$ with$_{\!}$ rotation$_{\!}$ equivariance,$_{\!}$ our$_{\!}$ algorithm$_{\!}$ naturally$_{\!}$ addresses$_{\!}$ the$_{\!}$ omnidi- rectional$_{\!}$ nature$_{\!}$ of$_{\!}$ fisheye$_{\!}$ images,$_{\!}$ yet$_{\!}$ using$_{\!}$ standard$_{\!}$ detec- tion$_{\!}$ network$_{\!}$ architecture$_{\!}$ designed$_{\!}$ for$_{\!}$ perspective$_{\!}$ images.

\vspace{-4pt}
\section{LOAF Dataset}
\vspace{-2pt}
\subsection{Dataset Acquisition}
\vspace{-2pt}
We first describe how our fisheye images are collected.

\noindent\textbf{Apparatus and Technical Specifications.} A TL-IPC59AE fisheye camera with 1.1mm focal length is adopted$_{\!}$ for~data recording.$_{\!}$ It$_{\!}$ has$_{\!}$ a$_{\!}$ wide$_{\!}$ FOV, reaching$_{\!}$ the$_{\!}$ full$_{\!}$ circle$_{\!}$ in$_{\!}$ the$_{\!}$ horizontal$_{\!}$ plane$_{\!}$ and$_{\!}$ 180$^{\circ}$ in$_{\!}$ the$_{\!}$ vertical$_{\!}$ plane.$_{\!}$ This$_{\!}$ offers a clear advantage in reducing deployment
cost -- installing just one fisheye camera instead of multiple conventional cameras to monitor the same region. However, the severe geometric distortions introduced prevent the use of standard detectors, which are designed for conventional cameras~\cite{cao2021handcrafted}.

\noindent\textbf{Data Capturing.} The fisheye camera is mounted on the ceiling (indoors) or poles (outdoors), 2.5$\sim$4.0 m from the ground with 200$\sim$300 m$^2$ FOV.  Considering the factors$_{\!}$ such$_{\!}$ as$_{\!}$ scenario$_{\!}$ diversity, pedestrians$_{\!}$ density$_{\!}$ and$_{\!}$ weather condition, we capture 110 fisheye image sequences in 80 realistic scenarios as the raw data pool. The recorded sequences span 14 hrs; the image resolution is 2952$\times$2952 pixels, and the frame rate is 10$\sim$20 fps. Eventually, 42,942 images, sampled at 1 fps, are collected to construct our LOAF dataset.

\begin{figure}
   \begin{center}
      \includegraphics[width=0.8\linewidth]{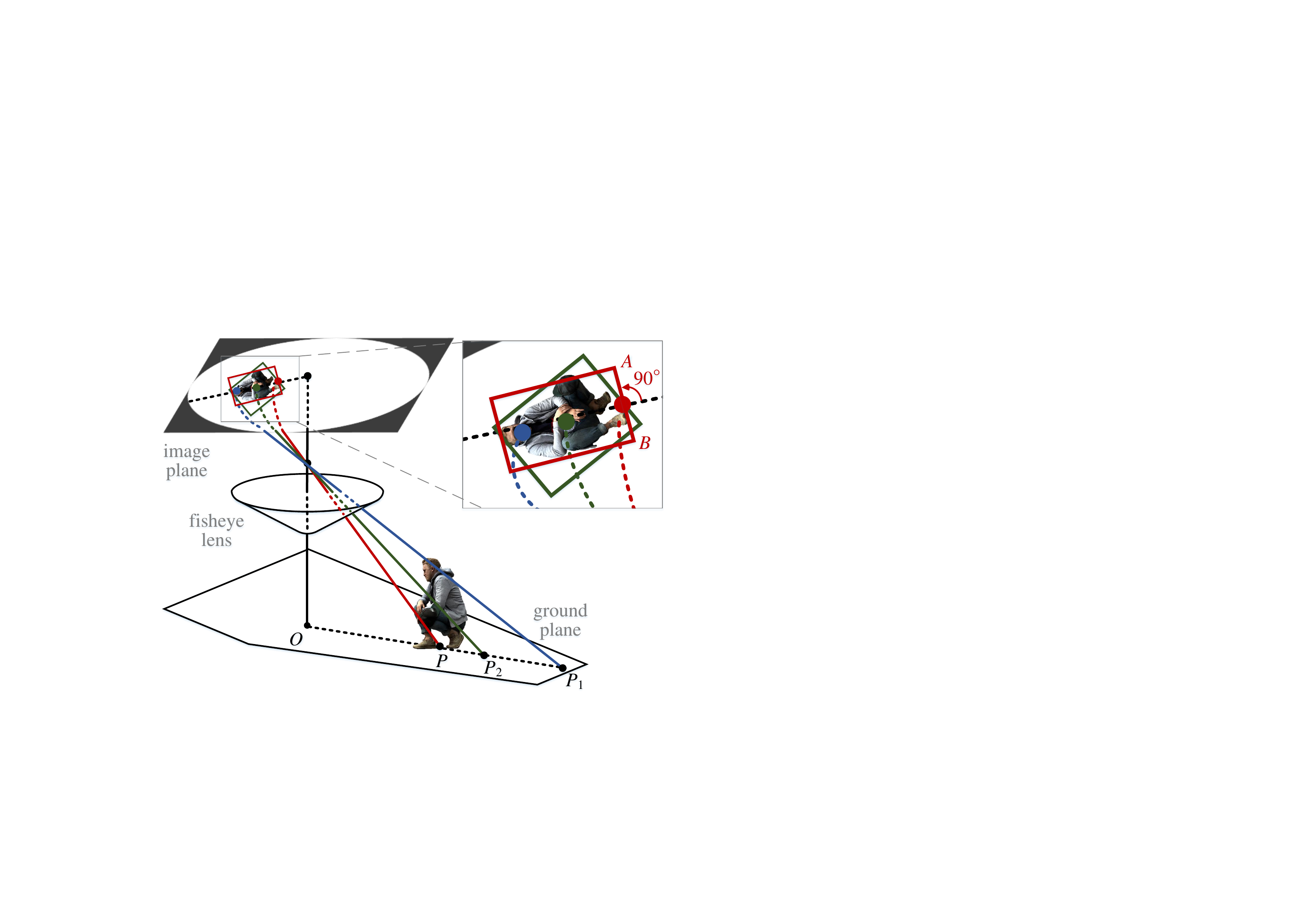}
   \end{center}
   \vspace{-14pt}
   \captionsetup{font=small}
   \caption{\small{\textbf{Human annotation comparison for visual positioning} (\S\ref{sec:dataset_annotation}). Previous methods typically use the detected human head~\cite{del2021robust} (\protect\includegraphics[scale=0.25,valign=c]{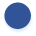}) or the center (\protect\includegraphics[scale=0.25,valign=c]{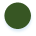}) of human-aligned body-box (\protect\includegraphics[scale=0.06,valign=c]{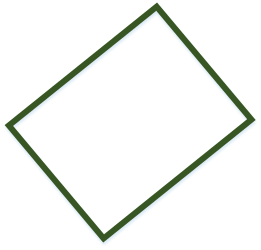})~\cite{li2019supervised,duan2020rapid} to determine the physical location $P$, bringing positioning errors (\ie, $PP_1$, $PP_2$). We instead leverage the radius-aligned box (\protect\includegraphics[scale=0.06,valign=c]{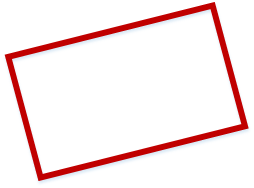}) as human representation. The midpoint (\protect\includegraphics[scale=0.25,valign=c]{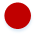}) of side $AB$, \ie, the closest point to the image center on the radius-aligned box, better corresponds to the human actual position.  }}
   \label{fig:annotation_comparison}
   \vspace{-14pt}
\end{figure}

\subsection{Dataset Annotation}\label{sec:dataset_annotation}
\vspace{-2pt}
We next describe how our fisheye images are annotated.

%%%%%%%%%%%%%%%%%%%%%%%%%%%%%Table 2%%%%%%%%%%%%%%%%%%%%%%%%%%%%%%%%%%%%%%%%%%%%
\begin{table*}[t]
\centering
\vspace{-5pt}
\begin{threeparttable}
\resizebox{0.99\textwidth}{!}{
\setlength\tabcolsep{2.5pt}
\renewcommand\arraystretch{1.02}
%\begin{tabular}{|c|l|r||c|c|c|c|c|c|ll|c|}  % {lccc}
\begin{tabular}{|r||c|c|c|c|c|c|c|c|c|c|c|c|c|}
\hline
%&\#
\multirow{2}{*}{Dataset~~~~~} &\multicolumn{2}{c|}{\#Scene} &\multirow{2}{*}{\#Video} &\multirow{2}{*}{\#Image} &\multicolumn{3}{c|}{\#People} &\multirow{2}{*}{Max Resolution} &\multirow{2}{*}{FOV (m$^2$)} &\multicolumn{3}{c|}{Annotation} &\multirow{2}{*}{FPS}\\%&C
\cline{2-3} \cline{6-8} \cline{11-13}
&Indoor &Outdoor & & &Total &Avg. &Max & & &People Detection &Location &Attribute &\\
\hline
\hline
%\multirow{5}{*}{\rotatebox{90}{Early}}
%&1
PIROPO~\cite{del2021robust} &2&0 &27&- &-&- &- &800$\times$600&$<$50 &Head Center &&&10\\%&\checkmark
HABBOF~\cite{li2019supervised} &2 &0 &4 &5,837 & 20,466&3.5 &5 &2,048$\times$2,048&$<$36 &Human-aligned box&&&12$\sim$30\\
MW-R~\cite{duan2020rapid}      & 6 & 0 &19 &8,752 & 22,825 & 2.6&6 & 1,488$\times$1,488&$<$36 & Human-aligned box&&&15\\
CEPDOF~\cite{duan2020rapid} &5&0 &8 &25,504 & 173,073&6.8 &13 &2,048$\times$2,048&$<$36 &Human-aligned box&&&1$\sim$10\\
WEPDTOF~\cite{tezcan2022wepdtof} &14&0&16 & 10,544 & 93,363& 8.9 & 35 & 2,592$\times$1,944 & $<$36 & Human-aligned box & & & 1$\sim$10\\
\hline
\textbf{LOAF} &11&39 &74 &42,942 &457,762 &10.5&65 &2,952$\times$2,952 &200$\sim$300 &Radius-aligned box&\checkmark&\checkmark&10$\sim$20\\
\hline
\end{tabular}
}
\end{threeparttable}
\captionsetup{font=small}
\caption{\small\textbf{\!Overhead$_{\!}$ fisheye$_{\!}$ datasets$_{\!}$ comparison}$_{\!}$ (\S\ref{sec:dataset_feature}). LOAF$_{\!}$ is the largest in terms of the total number of  pedestrian and scene categories.\!\!}
\label{table:dataset}
\vspace{-12pt}
\end{table*}

\noindent\textbf{Person Detection.} As the performance of localization relies critically on the quality of person detections, significant effort should be spent on~the fisheye person-detection annotation. Some human-detection representations were explored in previous datasets: \cite{del2021robust} labels the \textit{center point} of each human head; \cite{li2019supervised,duan2020rapid} opt for \textit{human-aligned} person-boxes. However, these representations have several shortcomings, especially with regards to the localization task. First, they suffer from some inherent limitations. The point-based representation is not applicable in other analysis tasks (\eg, re-identification, attribute recognition); the human-aligned box has ambiguity~\cite{chen2019fast,wang2019fast} -- the ground-truths of human-aligned boxes are not uniquely determined~\cite{duan2020rapid}. Second, and most important, these human-detection representations cannot meet the need of precise localization. Clearly, based on the fisheye camera model, one can project 2D detections on the 3D world for localization. However, the center of human head or of human-aligned box is not the exact placement occupied by human on the image plane, causing errors$_{\!}$ to$_{\!}$ estimations$_{\!}$ of$_{\!}$ human$_{\!}$ physical$_{\!}$ position$_{\!}$ (see$_{\!}$ Fig.$_{\!}$~\ref{fig:annotation_comparison}).$_{\!}$

We instead label each person through a \textit{radius-aligned} rectangular$_{\!}$ box.$_{\!}$ Such$_{\!}$ representation$_{\!}$ is$_{\!}$ favored$_{\!}$ as$_{\!}$ it:$_{\!}$ i)$_{\!}$ allows\! unique groundtruth box assignment; ii) fits well radially-oriented human bodies presented in fisheye images; and iii) better corresponds to the actual position of human on the image plane, facilitating physical localization.  Notably, although the radially-oriented box constraint was explored in a prior fisheye detector~\cite{li2019supervised}, there is no previous datasets provide such kind of annotation, neither is there any literature points out the advantage of such representation in localization. Finally, around 457K human box annotations are obtained. High-quality annotation is ensured via a rigorous quality check, conducted by highly skilled reviewers.

\begin{figure}[t]
   \begin{center}
      \includegraphics[width=0.9\linewidth]{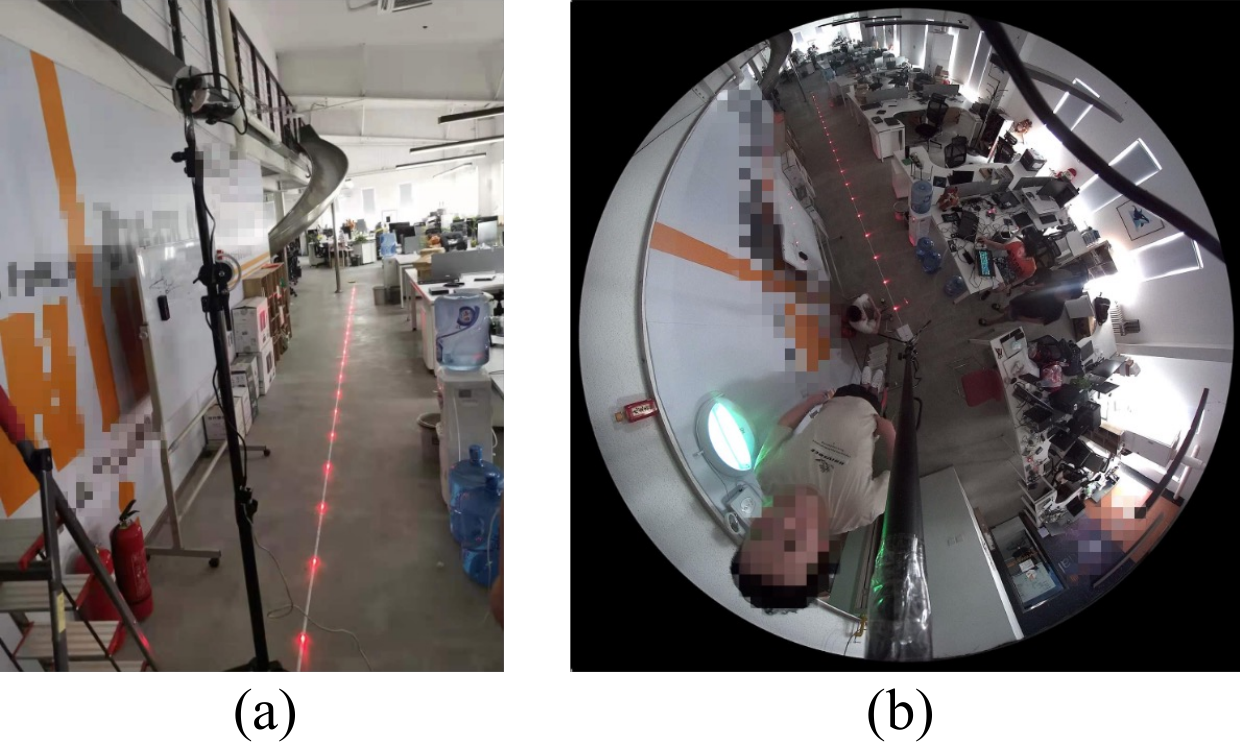}
   \end{center}
   \vspace{-17pt}
   \captionsetup{font=small}
   \caption{\small{\textbf{Calibration for groundtruth person location annotation.} (a) Ground makers
(with 0.05m precision). (b) Top-view fisheye image capture for calibration. See \S\ref{sec:dataset_annotation} for details.}}
   \label{fig:locationannotation}
   \vspace{-13pt}
\end{figure}

\begin{figure*}
   \begin{center}
   \vspace{-10pt}
      \includegraphics[width=\linewidth]{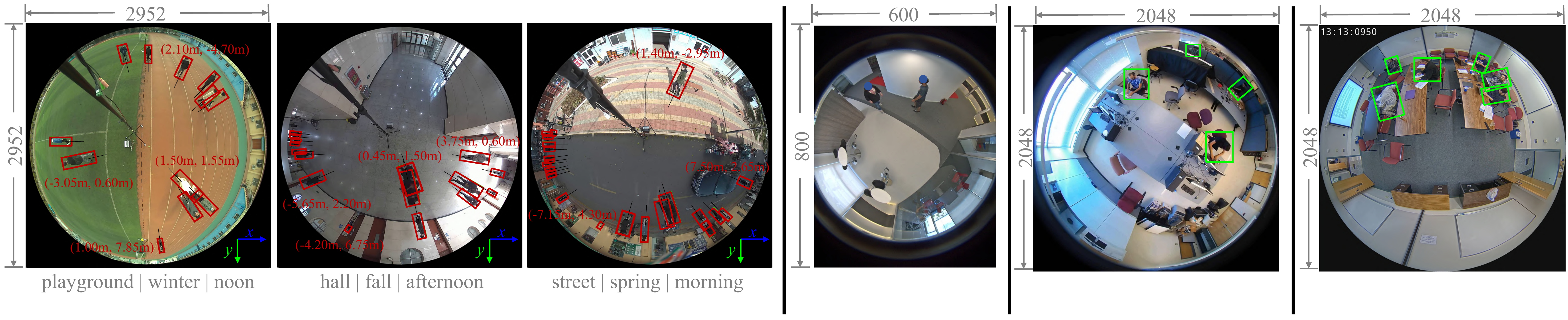}
      \put(-415,1){\small (a) Our proposed LOAF}
      \put(-243,1){\small (b) PIROPO~\cite{del2021robust}}
        \put(-165,1){\small (c) HABBOF~\cite{li2019supervised}}
        \put(-70,1){\small (d) CEPDOF~\cite{duan2020rapid}}
    \end{center}
   \vspace{-15pt}
   \captionsetup{font=small}
   \caption{\small{\textbf{Example images} from different datasets (\S\ref{sec:dataset_annotation}-\S\ref{sec:dataset_feature}). Prior datasets~\cite{del2021robust,li2019supervised,duan2020rapid} are restricted to few indoor scenes with person detection annotation only, \ie, head center (\protect\includegraphics[scale=0.25,valign=c]{figure/blue}) or human-aligned box (\protect\includegraphics[scale=0.3,valign=c]{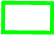}). In contrast, LOAF covers challenging indoor and outdoor scenes with human detection, localization and scene attribute annotations. The person-detection annotation is given as the radius-aligned box (\protect\includegraphics[scale=0.3,valign=c]{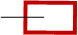}), which is more suitable for localization. For better visualization, we only present location ground-truths for some of persons. }}
   \label{fig:datasetsample}
   \vspace{-15pt}
\end{figure*}

\noindent\textbf{Person Localization.} For each scene, a 10\!~m ruler (with
0.05\!~m accuracy)$_{\!}$ is$_{\!}$ placed$_{\!}$ on$_{\!}$ the$_{\!}$ ground,$_{\!}$ and$_{\!}$ one$_{\!}$ end$_{\!}$ of$_{\!}$ the$_{\!}$ ruler is directly under the fisheye lens (\textit{cf}.\!~Fig.\!~\ref{fig:locationannotation}\!~(a)). We take a fisheye picture and use this marked picture as the calibration for all the dataset images recorded in this scene (\textit{cf}. Fig.~\ref{fig:locationannotation}\!~(b)). Finally, for each annotated human, the$_{\!}$ physical$_{\!}$ location,$_{\!}$ at$_{\!}$ sub-decimeter$_{\!}$ precision,$_{\!}$ is$_{\!}$ provided (\textit{cf}. Fig.~\ref{fig:datasetsample}).

\noindent\textbf{Scene Attribute.} To enable in-depth analysis, each image is annotated with multiple attributes, including \texttt{day}/\texttt{night}, \texttt{outdoor}/\texttt{indoor}, and \texttt{sunny}/\texttt{rain}/\texttt{foggy}/\texttt{snow}.

\vspace{-3pt}
\subsection{Dataset Design}\label{sec:dataset_design}
\vspace{-2pt}
We then list several key aspects of our dataset design.

\noindent\textbf{Privacy Protection.} To protect personal information, we apply the gaussian filter to blur all the visible facial regions in our dataset, and conduct experiments on the blurred data.

\noindent\textbf{Dataset Splits.}  LOAF contains 29,569 training, 4,600 validation, and 8,773 testing images (approximately 7 \texttt{train}, 1 \texttt{val}, and 2 \texttt{test}). Moreover, to better evaluate models' generalization ability, LOAF is split into five sets: \texttt{train} (7/28 indoor/outdoor scenes,  29,569 images), \texttt{val} \texttt{seen} (1/2 indoor/outdoor scenes, 1,700 images), \texttt{val} \texttt{unseen} (2/3 indoor/outdoor scenes, 2,900 images), \texttt{test} \texttt{seen} (2/3 indoor/outdoor scenes, 2,774 images), \texttt{test} \texttt{unseen} (2/8 indoor/outdoor scenes, 5,999 images). There are no overlapping scenes between \texttt{unseen} and \texttt{train} sets.

\noindent\textbf{Dataset$_{\!}$ Accessibility.}$_{\!}$ Only$_{\!}$ the$_{\!}$ desensitized$_{\!}$ version$_{\!}$ of$_{\!}$~our$_{\!}$ dataset$_{\!}$ will$_{\!}$ be$_{\!}$ released online, under$_{\!}$ the$_{\!}$ Creative Commons Attribution-NonCommercial-ShareAlike 4.0 License~\cite{CreativeCommons4.0}.

\begin{figure}
\captionsetup[subfloat]{captionskip=12pt}
    \centering
    \begin{tabular}{cc}
    %\hspace{-3pt}
    \adjustbox{valign=b}{\subfloat[Person scale]{%
          \vspace{5pt}
          \includegraphics[width=.40\linewidth]{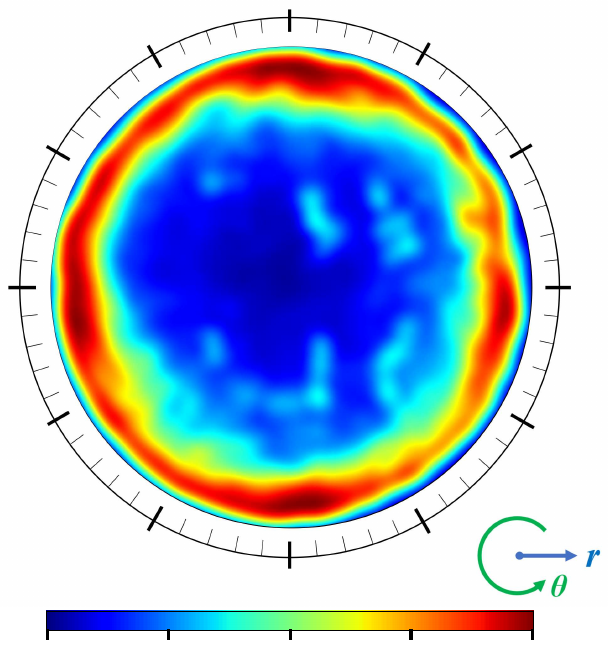}}}
      \put(-5.2,71.8){\scriptsize 0}
          \put(-11.3,95){\scriptsize 30}
          \put(-26.4,111.0){\scriptsize 60}
          \put(-52.5,118.8){\scriptsize 90}
          \put(-83.6,111.0){\scriptsize 120}
          \put(-99.1,95){\scriptsize 150}
          \put(-105.2,71.8){\scriptsize 180}
          \put(-98.1,47.6){\scriptsize 210}
          \put(-81.6,31.2){\scriptsize 240}
          \put(-54.6,25.5){\scriptsize 270}
          \put(-26.4,31.2){\scriptsize 300}
          \put(-11.8,47.6){\scriptsize 330}
          \put(-1.7,74.8){\tiny \scalebox{.72}{\degree}}
          \put(-4.3,97.5){\tiny \scalebox{.72}{\degree}}
          \put(-19.4,113.4){\tiny \scalebox{.72}{\degree}}
          \put(-45.5,121.8){\tiny \scalebox{.72}{\degree}}
          \put(-73.1,114){\tiny \scalebox{.72}{\degree}}
          \put(-88.2,97.8){\tiny \scalebox{.72}{\degree}}
          \put(-94.7,74.8){\tiny \scalebox{.72}{\degree}}
          \put(-87.9,50.4){\tiny \scalebox{.72}{\degree}}
          \put(-70.7,33.9){\tiny \scalebox{.72}{\degree}}
          \put(-44.0,28.5){\tiny \scalebox{.72}{\degree}}
          \put(-15.9,34.2){\tiny \scalebox{.72}{\degree}}
          \put(-1.3,50.6){\tiny \scalebox{.72}{\degree}}
          \put(-16.8,11.6){\scriptsize 480}
          \put(-36.3,11.6){\scriptsize 360}
          \put(-54.8,11.6){\scriptsize 240}
          \put(-73.8,11.6){\scriptsize 120}
          \put(-88.8,11.6){\scriptsize 0}
    &
    \hspace{12pt}
    \adjustbox{valign=b}{\begin{tabular}{@{}c@{}}
    \subfloat[Person locations]{%
          \includegraphics[width=.40\linewidth]{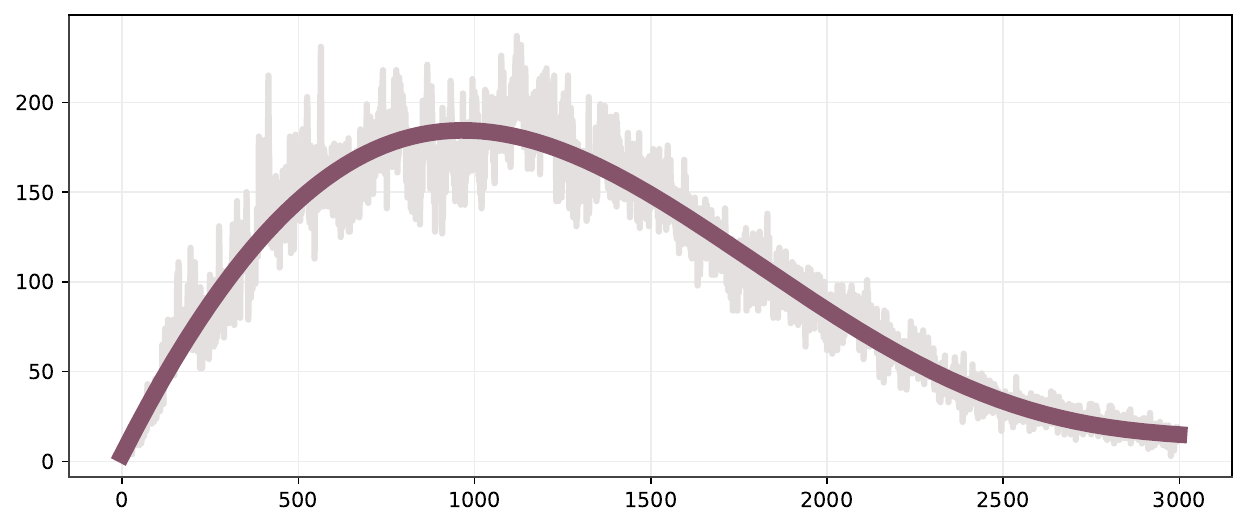}}
          \put(-91,-4.6){\tiny 0}
          \put(-78.,-4.6){\tiny 5k}
          \put(-64.5,-4.6){\tiny 10k}
          \put(-50.5,-4.6){\tiny 15k}
          \put(-36.5,-4.6){\tiny 20k}
          \put(-22.5,-4.6){\tiny 25k}
          \put(-8.5,-4.6){\tiny 30k}
          \put(-87,-10.8){\scriptsize Center Distance (centimeter)}
          \put(-98.5,1.0){\tiny 0}
          % \put(-119,12.7){\tiny 50}
          \put(-103.4,14.8){\tiny 100}
          % \put(-121.4,30){\tiny 150}
          \put(-103.4,29.6){\tiny 200}
          \put(-110,-6){\scriptsize \rotatebox{90}{Number of Boxes}}
          \vspace{-6pt} \\
    \subfloat[Person density]{%
          \includegraphics[width=.40\linewidth]{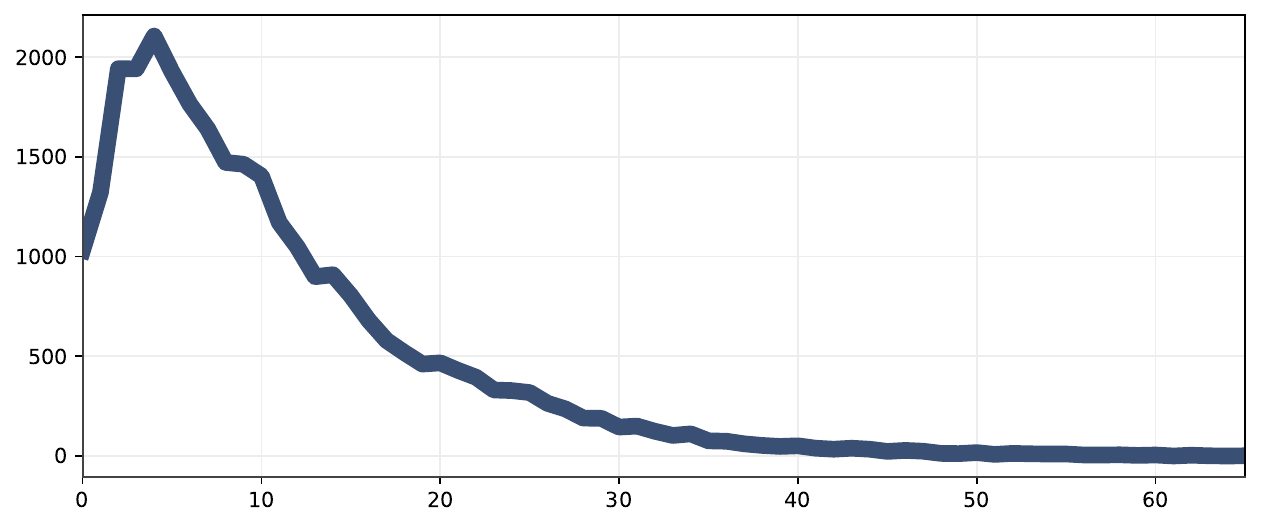}}
          \put(-83,-11){\scriptsize Number of Person Boxes}
          \put(-95,-5){\tiny 0}
          \put(-82,-5){\tiny 10}
          \put(-67.5,-5){\tiny 20}
          \put(-53.2,-5){\tiny 30}
          \put(-38.8,-5){\tiny 40}
          \put(-24.6,-5){\tiny 50}
          \put(-10.2,-5){\tiny 60}
          \put(-98.5,1.0){\tiny 0}
          % \put(-122.6,15.2){\tiny 0.5k}
          \put(-101.2,17.5){\tiny 1k}
          % \put(-122.6,34.4){\tiny 1.5k}
          \put(-101.2,33.4){\tiny 2k}
          \put(-110,-8){\scriptsize {\rotatebox{90}{Number of Images}}}
    \end{tabular}}
    \end{tabular}
    \vspace{-10pt}
   \captionsetup{font=small}
   \caption{\small{{\textbf{Dataset Statistics} (\S\ref{sec:dataset_feature}): We summarize LOAF with the distribution of \textbf{(a)} person scale (area in pixel), \textbf{(b)} person locations (horizontal distance between human and the fisheye lens in the real world), and \textbf{(c)} person density  (number of person per image).}}}
   \label{fig:statistics}
   \vspace{-15pt}
  \end{figure}
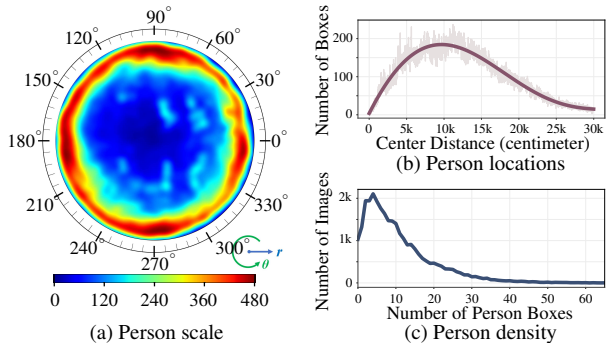

\vspace{-3pt}
\subsection{Dataset Features and Analysis}\label{sec:dataset_feature}
Finally,$_{\!~}$ we$_{\!~}$ present$_{\!~}$ statistic$_{\!~}$ analysis$_{\!~}$ of$_{\!~}$ LOAF$_{\!~}$ in$_{\!~}$ com-

\noindent parison with existing overhead fisheye datasets~\cite{del2021robust,li2019supervised,duan2020rapid}. LOAF distinguishes itself from three aspects (\textit{cf}. Table~\ref{table:dataset}):

\noindent\textbf{Large-scale}: LOAF has 42,942 fisheye images with more than 457K person boxes. Moreover, LOAF data are captured by an advanced fisheye camera, which is capable of covering a larger area (200$\sim$300 m$^2$) with higher pedestrian density (2$\sim$65 persons per scene, 10.5 in average). This makes LOAF the largest overhead fisheye dataset in terms of the total number of pedestrian and scene categories.

\noindent\textbf{High$_{\!}$ Diversity}: Existing$_{\!}$ datasets$_{\!}$ limit$_{\!}$ in$_{\!}$ data$_{\!}$ diversities, \ie, only$_{\!}$ containing$_{\!}$ very$_{\!}$ few$_{\!}$ indoor$_{\!}$ scenes$_{\!}$ (2$\sim$14)$_{\!}$ and$_{\!}$ completely$_{\!}$ missing$_{\!}$ outdoor$_{\!}$ scenarios$_{\!}$ (\textit{cf}.\!~Fig.\!~\ref{fig:datasetsample}). In contrast, LOAF$_{\!}$ involves$_{\!}$ 51$_{\!}$ realistic$_{\!}$ scenes, including$_{\!}$ 11$_{\!}$ indoor$_{\!}$ scenes (\eg, lab, office, library, classroom) and 40 outdoor scenes (\eg, street, playground, parking lot, square). The recorded data cover four seasons under different illumination (\eg, morning, noon, afternoon) and weather (\eg, sunny, rain, snow) conditions, and involve vast variance of human pose (\eg, walking, standing, and sitting), scale, location, and density (\textit{cf}. Fig.~\ref{fig:statistics}). Thus our dataset better reflects the distribution in real-world surveillance scenarios. 

\noindent\textbf{Rich and Positioning-aware Annotation}: LOAF is provided with rich ground-truths for detection, localization, and scene attribute, which lays a solid foundation for fisheye camera based human-centric analysis. Hence, as demonstr- ated in \S\ref{sec:dataset_annotation}, the radius-aligned human-box representation is adopted during our annotation. Compared with human-head center based point annotation \cite{del2021robust} and human-aligned person-boxes \cite{li2019supervised,duan2020rapid} used in previous datasets, radius-aligned human-boxes are more suitable for the position task.

%\noindent\textbf{Detailed Analysis.}

\vspace{-4pt}
\section{Our Approach}
\vspace{-2pt}
Our overhead fisheye camera based person localization solution comprises two parts. The former computes 2D detections that locate the people
on the image plane (\S\ref{sec:approach_stage1}). The latter converts the 2D detections to 3D-world coordinates, obtaining the physical location of the people  (\S\ref{sec:approach_stage2}).

\vspace{-4pt}
\subsection{Overhead Fisheye Person Detection}\label{sec:approach_stage1}
\vspace{-2pt}

\noindent\textbf{Core$_{\!}$ Idea:$_{\!}$ Rotation$_{\!}$ Equivariance.$_{\!}$} One of the reasons for the tremendous success of CNNs is their \textit{equivariance} to horizontal and vertical shifts and the resulting invariance to local deformations~\cite{lenc2015understanding,esteves2018learning}. This$_{\!}$ stimulates$_{\!}$ a$_{\!}$ line$_{\!}$ of$_{\!}$ efforts$_{\!}$ to$_{\!}$ learn$_{\!}$ robust$_{\!}$ representations$_{\!}$ equivariant to generic types of transformations$_{\!}$ \cite{hinton2011transforming,cohen2016group,wanglearning}. Formally,$_{\!}$ a$_{\!}$ representation$_{\!}$ $f_{\!}$ is$_{\!}$ said$_{\!}$ to$_{\!}$ be$_{\!}$ equivariant$_{\!}$ with$_{\!}$ a$_{\!}$ geometric$_{\!}$ transformation$_{\!}$ $g$$_{\!}$ (\eg,$_{\!}$ cropping, flipping) for an input (image) $I$ if:
\vspace{-4pt}
\begin{equation}\small
% \vspace{-4pt}
\begin{aligned}
f(g({I}))\approx g(f(I)).
\end{aligned}
\label{equ:rotate}
\vspace{-4pt}
\end{equation}
That is to say, the output representation $f(I)$ is changed in the$_{\!}$ same$_{\!}$ way$_{\!}$ as$_{\!}$ the$_{\!}$ transformation$_{\!}$ $g_{\!}$ imposed$_{\!}$ to$_{\!}$ the$_{\!}$ input$_{\!}$ $I$.$_{\!}$

In$_{\!}$ this$_{\!}$ work,$_{\!}$ we$_{\!}$ devise$_{\!}$ a$_{\!}$ \textit{query}-based$_{\!}$ fisheye$_{\!}$ person$_{\!}$ de-  tector$_{\!}$ that$_{\!}$ exhibits$_{\!}$ 360$^{\circ}$-rotational$_{\!}$ equivariance$_{\!}$ through$_{\!}$ an elaborately$_{\!}$ designed$_{\!}$ training$_{\!}$ protocol.$_{\!}$ Our key insight is derived from the omnidirectional nature of this task: if one rotates the input fisheye image by an arbitrary angle, then the outputs of a robust fisheye detector should change accord- ingly.$_{\!}$ We$_{\!}$ thus$_{\!}$ design$_{\!}$ a$_{\!}$ \textit{rotation}$_{\!}$ \textit{equivariant}$_{\!}$ training$_{\!}$ strategy, which forces the matching between object queries and image representations to be equivariant against 360$^{\circ}$-rotations. In$_{\!}$ this$_{\!}$ way,$_{\!}$ the$_{\!}$ rotational$_{\!}$ symmetry$_{\!}$ of$_{\!}$ omnidirectional$_{\!}$ im- ages$_{\!}$ is$_{\!}$ explicitly$_{\!}$ addressed,$_{\!}$ without$_{\!}$ architectural$_{\!}$ modification of standard detectors developed for perspective images.

\begin{figure}
   \vspace{-6pt}
   \begin{center}
      \includegraphics[width=0.99\linewidth]{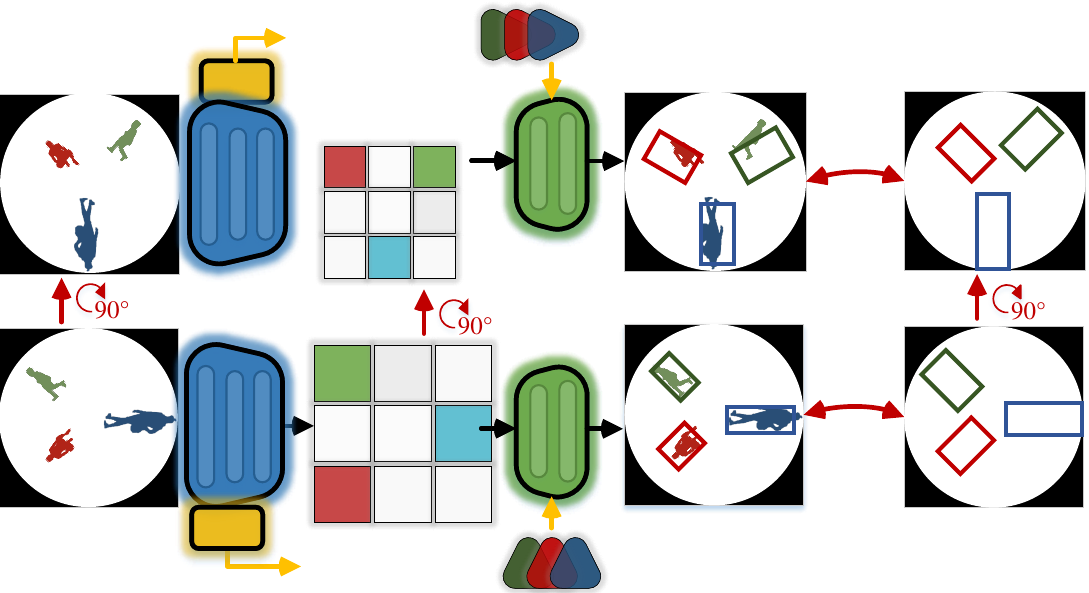}
            \put(-232,112){\small $g^{r\!}({I})$}
      \put(-232,60){\small $I$}
      \put(-186,108){\small $\kappa$}
      \put(-187,84){\small $\mathcal{F}$}
      \put(-187,33){\small $\mathcal{F}$}
      \put(-188,12){\small $\kappa$}
      \put(-168,116){\small $\{\bm{q}_n^{g^r\!}\}_{n=1}^N$}
      \put(-160,102){\small $g^r(\bm{I})$}
      \put(-164,58){\small $\bm{I}$}
      \put(-164,3){\small $\{\bm{q}_n\}_{n=1}^N$}
      \put(-119,32){\small $\mathcal{D}$}
      \put(-119,89){\small $\mathcal{D}$}
      \put(-57,44){\tiny $\mathcal{L}_\text{det}$}
      \put(-60.5,95){\tiny \scalebox{.85}{$\mathcal{L}_\text{rotat-equi}$}}
      \put(-41,10){\small $\{\bm{b}_{\ell(n)}\}_{n=1}^{{N}}$}
      \put(-41,112){\small $\{\bm{b}^{g^r}_{\ell(n)}\}_{n=1}^{{N}}  $}
      %\put(-205,27){\footnotesize $(u,v)$}
%     \put(-135,41){\footnotesize $(\!\Delta\!u\!,\!\Delta\!v\!)$}
   \end{center}
   \vspace{-14pt}
   \captionsetup{font=small}
   \caption{\small{\textbf{Our rotation equivariant training strategy} for query-based$_{\!}$ fisheye$_{\!}$
person$_{\!}$ detection$_{\!}$ (\S\ref{sec:approach_stage1}).$_{\!}$ For$_{\!}$ equivariant$_{\!}$ training,$_{\!}$ object queries $\{\bm{q}_n^{g^r\!}\}_{n=1}^N$ created for rotated image $g^{r\!}(I)$ and
the rotated representation $g^{r\!}(\bm{I})$ are fed into the decoder $\mathcal{D}$; the rotated ground-truths $\{\bm{b}^{g^r}_{\ell(n)}\}_{n=1}^{{N}}$ are set as the training targets. }}
   \label{fig:training_method}
   \vspace{-8pt}
\end{figure}
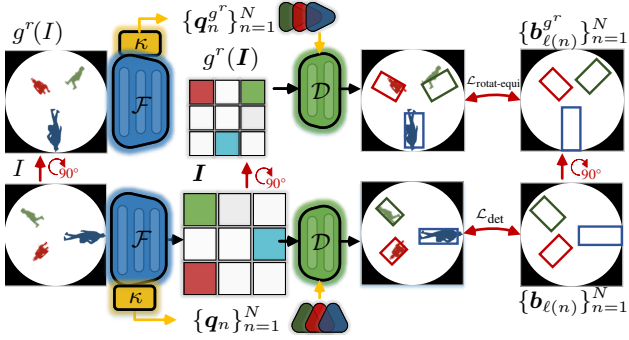

\noindent\textbf{Rotation$_{\!}$ Equivariant$_{\!}$ Training$_{\!}$ for$_{\!}$ Query-based$_{\!}$ Fisheye$_{\!}$ Detection.$_{\!}$} As$_{\!}$ shown$_{\!}$ in$_{\!}$ Fig.$_{\!}$~\ref{fig:training_method},$_{\!}$ our$_{\!}$ fisheye$_{\!}$ detector$_{\!}$ is$_{\!}$ built$_{\!}$ upon$_{\!}$ DETR$_{\!}$~\cite{carion2020end},$_{\!}$ promoted$_{\!}$ by$_{\!}$ a$_{\!}$ rotation$_{\!}$ equivariant$_{\!}$ training$_{\!}$ strategy.$_{\!~}$ Basically,$_{\!~}$ DETR$_{\!~}$ conducts$_{\!~}$ detection$_{\!~}$ in$_{\!~}$ a$_{\!~}$ query-

\noindent based fashion.$_{\!}$ Denote$_{\!}$ $\mathcal{F}_{\!}$ as$_{\!}$ a$_{\!}$ \textit{feature$_{\!}$ encoder}$_{\!}$~\includegraphics[scale=0.12,valign=c]{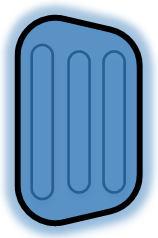}$_{\!}$ that$_{\!}$ extracts representation$_{\!}$~\includegraphics[scale=0.1,valign=c]{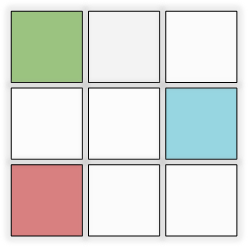}$_{\!}$ $\bm{I}_{\!}$ of$_{\!}$ image$_{\!}$ $I$,$_{\!}$ and$_{\!}$ $\kappa_{\!}$ as$_{\!}$ a$_{\!}$ \textit{query$_{\!}$ creator}$_{\!}$~\includegraphics[scale=0.25,valign=c]{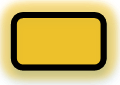} that$_{\!}$~outputs$_{\!}$ a$_{\!}$ set$_{\!}$ of$_{\!}$ $N_{\!}$ object-aware$_{\!}$ descriptors$_{\!}$~\includegraphics[scale=0.25,valign=c]{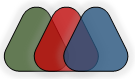}$_{\!}$ $\{\bm{q}_n\}_{n=1}^N$ from $\bm{I}$. DETR employs $\{\bm{q}_n\}_{n=1}^N$ as queries to retrieve target objects from $\bm{I}$:
\vspace{-3pt}
\begin{equation}\small
% \vspace{-4pt}
\begin{aligned}
\{\hat{\bm{b}}_n\}_{n=1}^{{N}} = \mathcal{D}(\bm{I}, \{\bm{q}_n\}_{n=1}^N),
\end{aligned}
\label{equ:decoding}
\vspace{-0pt}
\end{equation}
{where}$_{\!}$ $\bm{I}\!=\!\mathcal{F}(I)$,$_{\!}$ $\{\bm{q}_n\}_{n=1\!}^N\!=\!\kappa(\bm{I})$,$_{\!}$ and$_{\!}$ $\mathcal{D}_{\!}$ refers$_{\!}$ to$_{\!}$ the$_{\!}$~Trans-
 former$_{\!}$ \textit{decoder}$_{\!}$~\includegraphics[scale=0.12,valign=c]{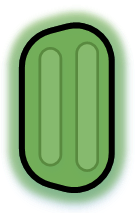}$_{\!}$ which$_{\!}$ formulates$_{\!}$ the$_{\!}$ object-query$_{\!}$ match- ing$_{\!}$ process$_{\!}$ via$_{\!}$ neural$_{\!}$ cross-attention$_{\!}$ computation.$_{\!}$ $\{\hat{\bm{b}}_n\}_{n=1}^{{N}}$ are$_{\!}$ the$_{\!}$ set$_{\!}$ of$_{\!}$ predicted$_{\!}$ parameterized$_{\!}$ object$_{\!}$ bounding$_{\!}$ boxes.$_{\!}$

Our rotation equivariant training further encourages robust object-query matching which is equivariant against ro- tations.$_{\!}$ First,$_{\!}$ given$_{\!}$ an$_{\!}$ input$_{\!}$ rotation$_{\!}$ transformation$_{\!}$ $g^r$,$_{\!}$ a$_{\!}$ ro- bust fisheye detector is desired to be able to extract rotation-equivariant representation. Analogous to Eq.~\ref{equ:rotate}, we have:
\vspace{-4pt}
\begin{equation}\small
% \vspace{-4pt}
\begin{aligned}
\mathcal{F}(g^r({I}))\approx g(\mathcal{F}(I))=g^r(\bm{I}).
\end{aligned}
\label{equ:rotate1}
\vspace{-5pt}
\end{equation}

Also, it is reasonable to assume that the final output of a robust fisheye detector should be changed in the same way to the rotation $g^r$ applied to the input fisheye image $I$:
\vspace{-2pt}
\begin{equation}\small
\begin{aligned}
\{\bm{b}^{g^r}_{\ell(n)}\}_{n=1}^{{N}} \approx \mathcal{D}(\bm{I}^{g^r}, \{\bm{q}_n^{g^r}\}_{n=1}^N),
\end{aligned}
\label{equ:rotate2}
\vspace{-1pt}
\end{equation}
where $\bm{I}^{g^r\!\!}$ indicates$_{\!}$ the$_{\!}$ feature$_{\!}$ of$_{\!}$ rotated$_{\!}$ image$_{\!}$ $g^r(I)$,$_{\!}$~\ie, $\bm{I}^{g^r\!\!\!}\!=_{\!}\!\mathcal{F}(g^{r\!}({I}))$.$_{\!}$ Similarly,$_{\!}$ $\{\bm{q}_n^{g^r\!}\}_{n=1\!\!}^N$ denote$_{\!}$ the$_{\!}$ object$_{\!}$ queries$_{\!}$ \includegraphics[scale=0.25,valign=c]{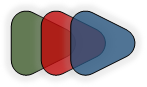}$_{\!}$ derived$_{\!}$ from$_{\!}$ $g^{r\!}(I)$,$_{\!}$ \ie,$_{\!}$ $\{\bm{q}_n^{g^r\!}\}_{n=1\!}^N\!=\!\!\kappa(\bm{I}^{g^r\!})$.$_{\!}$ $\{\bm{b}^{g^r}_{\ell(n)}\}_{n=1}^{{N}}$ are the rotated groundtruth bounding boxes, where $\ell(n)$ returns the groundtruth index for $n$-th query.

Considering Eq.\!~\ref{equ:rotate1} and Eq.\!~\ref{equ:rotate2}, we can have: $\{\bm{b}^{g^r}_{\ell(n)}\}_{n=1}^{{N}} \approx \mathcal{D}(g^{r\!}(\bm{I}), \{\bm{q}_n^{g^r\!}\}_{n=1}^N)$.$_{\!}$ Hence$_{\!}$ our$_{\!}$ rotation$_{\!}$ equivariant$_{\!}$ training$_{\!}$ objective$_{\!}$ is given as:
\vspace{-2pt}
\begin{equation}\small
\begin{aligned}
 \mathcal{L}_\text{rotat-equi} = \mathcal{L}_\text{det}\big(\{\bm{b}^{g^r}_{\ell(n)}\}_{n=1}^{{N}}, \mathcal{D}(g^{r\!}(\bm{I}), \{\bm{q}_n^{g^r\!}\}_{n=1}^N)\big).
\end{aligned}
\label{equ:loss_equiv}
\vspace{-1pt}
\end{equation}
Here$_{\!}$ $\mathcal{L}_\text{det\!}$ is$_{\!}$ the$_{\!}$ standard$_{\!}$ detection$_{\!}$ loss$_{\!}$ in$_{\!}$ DETR$_{\!}$~\cite{carion2020end}.$_{\!}$ As$_{\!}$ such,$_{\!}$ the rotation equivariance properties for both fisheye representation $\mathcal{F}$ (\textit{cf}. Eq.\!~\ref{equ:rotate1}) and the query-based detection prediction $\mathcal{D}(\bm{I}, \{\bm{q}_n\}_{n=1}^N)$ (\textit{cf}. Eq.\!~\ref{equ:rotate2}) are sought in a single training target. This also allows us to effortlessly adapt standard DETR$_{\!}$ to$_{\!}$ our$_{\!}$ omnidirectional$_{\!}$ detection$_{\!}$ setting.$_{\!}$ Note$_{\!}$ that$_{\!}$~our rotation$_{\!}$ equivariant$_{\!}$ training$_{\!}$ differs$_{\!}$ from$_{\!}$ rotational$_{\!}$ data$_{\!}$ aug- mentation$_{\!}$ technique$_{\!}$ which$_{\!}$ only$_{\!}$ views$_{\!}$ rotated$_{\!}$ images$_{\!}$ as$_{\!}$ in- dividual$_{\!}$ training$_{\!}$ samples$_{\!}$ (see$_{\!}$ \S\ref{sec:exabs}$_{\!}$ for$_{\!}$ detailed$_{\!}$ experiments).$_{\!}$

\subsection{2D-3D Projection based Person Localization}\label{sec:approach_stage2}
\begin{SCfigure}[0.58]
      \includegraphics[width=0.60\linewidth]{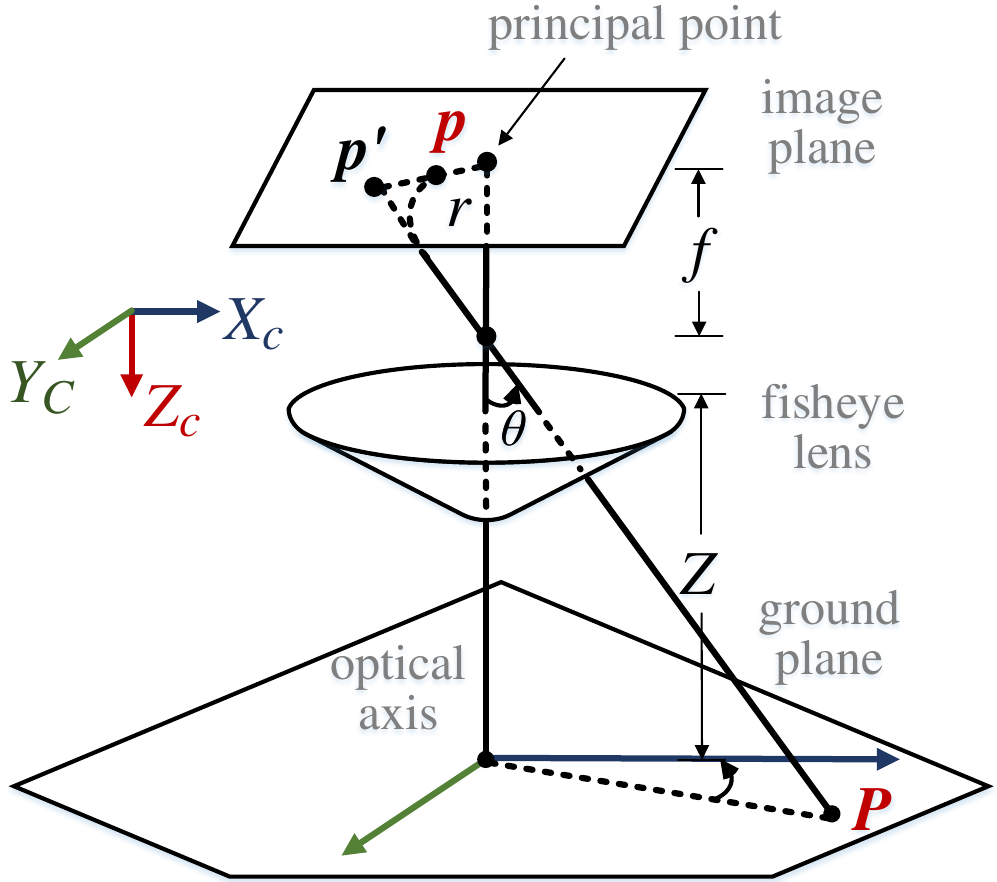}
        \put(-34,12){\fontsize{7pt}{6.2pt}\selectfont  {$\phi$}}
   \hspace*{-5pt}
   \captionsetup{font=small}
   \caption{\small{\textbf{$_{\!}$Our~fisheye camera model$_{\!}$ based person$_{\!\!}$ localization} (\S\ref{sec:approach_stage2}).$_{\!}$ Here$_{\!}$~$\bm{p}_{\!}$ and$_{\!}$~$\bm{p}'_{\!}$ denote$_{\!}$ the$_{\!}$ projection$_{\!}$ point$_{\!}$ of$_{\!}$ the distortion$_{\!}$ point$_{\!}$ $\bm{P}_{\!}$ by$_{\!}$ fisheye$_{\!}$ lens and$_{\!}$ normal$_{\!}$ perspective$_{\!}$ lens, respectively.}}
   \label{fig:localization}
   \vspace{-6pt}
\end{SCfigure}

\noindent\textbf{General Fisheye Camera Model.} The perspective projection of a normal pinhole camera can be written as
$r=f\tan\theta$, where $r$ indicates the projection distance between
the principal point and the image point, $f$ is the focal length, and $\theta$ is
the angle between the incident ray and the camera's optical axis. However, fisheye lens does not follow this perspective projection model~\cite{miyamoto1964fish}, as the FOV equals to 180$^{\circ}$ (\textit{cf}. Fig.~\ref{fig:localization}). Fortunately, the image formation of different kinds of fisheye lenses can be approximated by a general polynomial projection model~\cite{kannala2006generic}, \ie,
\begin{equation}\small
\begin{aligned}
r(\theta) = \sum^n\nolimits_{i=1}k_i\theta^{2i-1},~~~~n=1, 2, 3, 4, \cdots
\end{aligned}
\label{equ:fisheyemodel}
\end{equation}
High distortions can be handled well when $n=5$~\cite{kannala2006generic}. The coefficients $k$s can be obtained from camera calibration.

\begin{table*}[t]
   \centering
     \vspace{-4pt}
   \resizebox{0.99\textwidth}{!}{
   \setlength\tabcolsep{1pt}
   \renewcommand\arraystretch{1.03}
   \begin{tabular}{|c ||ccc|ccc|cc ||ccc|ccc|cc||c|}
   \hline
   % \thickhline
   % \rowcolor{mygray}
   & \multicolumn{8}{c||}{\texttt{val}} & \multicolumn{8}{c|}{\texttt{test}} & \\
   \cline{2-17}
   % \rowcolor{mygray}
   \multirow{-2}{*}{Method} & mAP\! $\uparrow$ &  AP$_{50}$$\uparrow$ & AP$_{75}$$\uparrow$ &AP$_{n}$$\uparrow$& AP$_{m}$$\uparrow$ & AP$_{f}$$\uparrow$ &  AP$_{seen}$$\uparrow$ & AP$_{unseen}$$\uparrow$ & mAP\! $\uparrow$ &  AP$_{50}$$\uparrow$& AP$_{75}$$\uparrow$ &AP$_{n}$$\uparrow$& AP$_{m}$$\uparrow$ & AP$_{f}$$\uparrow$ &  AP$_{seen}$$\uparrow$ & AP$_{unseen}$$\uparrow$  & \multirow{-2}{*}{FPS\! $\uparrow$} \\
   \hline
   \hline
   Seide \etal~\cite{seidel2019improved} & 21.8  & 59.8 & 7.6  &32.8 &28.5 & 2.3 & 23.2 &19.5
   & 20.2 &58.2 & 7.1& 32.2&28.3 &3.9 & 22.4& 18.9&10.2\\
    Li \etal~\cite{li2019supervised} & 28.5  & 63.3  & 20.1 & 46.8& 24.2& 1.3  & 33.8 & 29.3
    & 27.2 & 65.2& 21.3&47.2 &24.8 & 1.3& 31.8&27.6 &0.6\\
    Tamura \etal~\cite{tamura2019omnidirectional} & 34.8 & 72.1 & 27.7 &51.7 &38.8 & 8.7 & 38.8 &32.9
    & 34.2& 72.8& 28.7& 53.7& 37.3& 6.5& 39.5& 33.2&10.2\\
    RAPiD~\cite{duan2020rapid} & 40.3 & 77.9 & 34.8 & 55.3& 41.9& 9.2  & 44.7 &37.6
    &39.2 &77.9 &35.4 &54.8 &40.1 &7.9 &44.2 & 37.3&8.4\\ \hline
    Ours &  \textbf{47.2} & \textbf{82.3} & \textbf{48.2} &  \textbf{63.8} & \textbf{54.1} &\textbf{14.0}& \textbf{50.6} & \textbf{45.5}
    & \textbf{46.2}&\textbf{81.1} & \textbf{47.3}& \textbf{66.1}&\textbf{53.5} & \textbf{12.6} & \textbf{49.3}& \textbf{44.9} &\textbf{12.1}\\
   %\hline
   \hline
   \end{tabular}
   }
   \captionsetup{font=small}
   \caption{\small{\textbf{Person detection results} on \texttt{val} and \texttt{test} sets of LOAF (\S\ref{sec:exPD}). }}
    \label{table:loaf_det}
   \vspace{-12pt}
\end{table*}

 % \hspace*{20pt}

\noindent\textbf{2D-3D Projection.} Given a detected human location point $\bm{p}\!=\!(u, v)^{\top}$ in the fisheye image pixel coordinate system, our target is to calculate its 3D location $\bm{P} = (X, Y, Z)^{\top}$ in the camera coordinate system, where $Z$ is the altitude of the fisheye camera and priorly known. To do so, one can first calculate $r$ as: $r\!=\!\sqrt{(x^2+y^2)}$, where $x\!=\!(u\!-\!u_0)/f$ and $y\!=\!(v\!-\!v_0)/f$, and $(u_0,v_0)^{\top}$ are the coordinates of the principal point in the fisheye image. Then $\theta$ can be obtained by solving Eq.~\ref{equ:fisheyemodel} using a numerical means~\cite{kannala2006generic}. With the altitude $Z$ of the camera, the actual location $(X,Y)^{\top\!}$ is given as:
   \vspace{-2pt}
\begin{equation}\small
\begin{aligned}
\binom{X}{Y}=Z\binom{\tan\theta\cos\phi}{\tan\theta\sin\phi},
\end{aligned}
\label{equ:location}
\end{equation}
where $\phi=\arctan(y,x)$ refers to the polar angle, which is shared by both $\bm{p}$ and $\bm{P}$.

It can be seen that precise determination of the person location point $\bm{p}$ on the image plane is vital for estimating the corresponding physical position $\bm{P}$. However, most previous approaches, restricted to axis-aligned~\cite{zhu2019object,seidel2019improved} or person-aligned~\cite{duan2020rapid} human-box representation, use the center of the detection box to approximate $\bm{p}$. Few exceptions~\cite{del2021robust}, built on point-based human representation, treat the center of the detected human head as $\bm{p}$. However, they suffer from a similar issue as the human head center is even often far from the exact position $\bm{p}$ human stand on, \wrt the image plane. Instead, with our radius-aligned human-box representation, the closest point to the principal point on the bounding box better corresponds to $\bm{p}$.

\subsection{Implementation Details}

\noindent\textbf{Network Architecture.} Our fisheye person detector is built upon DAB-DETR$_{\!}$~\cite{liu2022dab}, a prevalent variant of DETR$_{\!}$~\cite{carion2020end} but converges much faster. Swin-T~\cite{liu2021swin} is$_{\!}$ utilized$_{\!}$ as$_{\!}$ the$_{\!}$ backbone. For our LOAF with radius-aligned person boxes, the output of our detector is a 4D vector $\hat{\bm{b}}\!\in\![0, 1]^4$ that parameterizes 2D center coordinates, height, and width. Note that, for traditional fisheye detection datasets like~\cite{duan2020rapid,tezcan2022wepdtof} with arbitrary-oriented person box annotations, an extra output dimension is needed for rotation angle regression.

\noindent\textbf{Training Objective.} Our fisheye person detector is end-to-\\
\noindent end$_{\!}$ trained$_{\!}$ by$_{\!}$ jointly$_{\!}$ optimizing$_{\!}$ the$_{\!}$ vanilla$_{\!}$ detection$_{\!}$ loss used in DETR~\cite{carion2020end,liu2022dab} (referred as $\mathcal{L}_\text{Det}$) and our proposed rotation-equivariant constraint (\ie, $\mathcal{L}_\text{rotat-equi}$ in Eq.~\ref{equ:loss_equiv}):
\vspace{-3pt}
\begin{equation}\small
\begin{aligned}
 \mathcal{L} = \mathcal{L}_\text{det} + \lambda\mathcal{L}_\text{rotat-equi},
\end{aligned}
\label{equ:loss_all}
\vspace{-3pt}
\end{equation}
where the coefficient $\lambda$ is empirically set to 0.5.

\begin{table}[t]
   \centering
   \resizebox{0.49\textwidth}{!}{
   \setlength\tabcolsep{2pt}
   \renewcommand\arraystretch{1.03}
   \begin{tabular}{|c||ccc|ccc|}
   \hline
   Method & mAP\! $\uparrow$& AP$_{50}$\! $\uparrow$ & AP$_{75}$\! $\uparrow$  & Precision\! $\uparrow$  & Recall\! $\uparrow$ & F-Score\! $\uparrow$ \\
   \hline
   \hline
    Seide \etal~\cite{seidel2019improved} &20.9& 50.6 & 10.2 & 80.6 & 39.5& 53.0 \\
     Li \etal~\cite{li2019supervised}  & 34.2& 75.7 & 28.6& 86.3  & 65.4 & 74.4 \\
     Tamura \etal~\cite{tamura2019omnidirectional} & 29.3& 61.0 & 23.4&88.8&51.2&65.0 \\
     RAPiD~\cite{duan2020rapid}  & 39.3& 85.4 & 26.0& 89.2 &78.7 & 83.6 \\ \hline
     Ours & \textbf{46.8}  & \textbf{88.1}  & \textbf{36.8} & \textbf{90.2} & \textbf{87.4} &  \textbf{88.6}\\
    \hline
   \end{tabular}
   }
   \captionsetup{font=small}
   \caption{\small\textbf{Person detection results} on CEPDOF~\cite{duan2020rapid} (\S\ref{sec:exPD}).}
   \label{table:fish_det1}
   \vspace{-13pt}
\end{table}

\noindent\textbf{Training.}$_{\!}$ Our$_{\!}$ fisheye$_{\!}$ detector$_{\!}$ is$_{\!}$ trained$_{\!}$ with$_{\!}$ a$_{\!}$ batch$_{\!}$ size$_{\!}$ of$_{\!}$ 8$_{\!}$ for$_{\!}$ 50$_{\!}$ epochs,$_{\!}$ where$_{\!}$ the$_{\!}$ AdamW\!~\cite{loshchilovdecoupled}$_{\!}$ optimizer$_{\!}$  is$_{\!}$ employed$_{\!}$ with$_{\!}$ base learning$_{\!}$ rate$_{\!}$ 2e-4$_{\!}$ and$_{\!}$ decayed$_{\!}$ by$_{\!}$ 0.1$_{\!}$ at$_{\!}$ epoch$_{\!}$ 40. The remaining hyper-parameters are determined$_{\!}$ following$_{\!}$ \cite{tamura2019omnidirectional,duan2020rapid}.$_{\!}$ Specifically,$_{\!}$ we$_{\!}$ initialize$_{\!}$ backbones$_{\!}$ with$_{\!}$ Image- Net\!~\cite{deng2009imagenet}$_{\!}$ pre-trained weights and adopt standard data augmentation  techniques, \ie, color jitter, horizontal flip, and random scaling, with a base training size of$_{\!}$ 608$\times$608.$_{\!}$ For$_{\!}$ CEPDOF$_{\!}$~\cite{duan2020rapid}$_{\!}$ and$_{\!}$ WEPDTOF$_{\!}$~\cite{tezcan2022wepdtof},$_{\!}$ we$_{\!}$ use$_{\!}$ a$_{\!}$ batch$_{\!}$ size$_{\!}$ of$_{\!}$ 128$_{\!}$ and pre-train the detector on COCO~\cite{lin2014microsoft} for 50 epochs to prevent over-fitting, as in$_{\!}$~\cite{duan2020rapid,tezcan2022wepdtof}. For the computation of our rotation-equivariant loss $\mathcal{L}_\text{rotat-equi}$, the training images are rotated by a degree randomly sampled from 0 to 360.  Our$_{\!}$ detector$_{\!}$ is$_{\!}$ implemented$_{\!}$ in$_{\!}$ PyTorch$_{\!}$ and$_{\!}$ trained$_{\!}$ on$_{\!}$ eight 

\noindent NVIDIA Tesla V100 GPUs with a 32GB memory per-card.

\noindent\textbf{Inference.}$_{\!}$ Once$_{\!}$ trained,$_{\!}$ our$_{\!}$ fisheye$_{\!}$ detector$_{\!}$ can$_{\!}$ be$_{\!}$ directly applied for locating persons on the omnidirectional image plane. After that, the physical locations of the detected persons can be easily obtained through the numerical solution described in \S\ref{sec:approach_stage2}. For fisheye person detection, we follow prior work~\cite{seidel2019improved,li2019supervised,tamura2019omnidirectional,duan2020rapid} to use 1024$\times$1024 input resolution without any test-time augmentation or post-processing. Testing is conducted on a single NVIDIA V100 GPU with 16 GB memory.

\vspace{-2pt}
\section{Experiment}
\vspace{-2pt}
\subsection{Experimental Setup}
\vspace{-2pt}
\noindent\textbf{Evaluation Protocol.} On the top of LOAF, we conduct~ex- periments for fisheye based person detection and localization. Following the dataset splitting~(\textit{cf}. \S\ref{sec:dataset_design}), performance are reported on  \texttt{seen} and \texttt{unseen} scenes, respectively, for both \texttt{val} and  \texttt{test} sets. This allows us to assess the generalization ability over different surveillance scenarios. For comprehensive study, we further report person detection performance on two existing fisheye datasets, CEPDOF\!~\cite{duan2020rapid} and WEPDTOF\!~\cite{tezcan2022wepdtof}. Note that localization cannot be test since$_{\!}$ \cite{duan2020rapid,tezcan2022wepdtof}$_{\!}$ only$_{\!}$ provide$_{\!}$ person$_{\!}$ bounding$_{\!}$ box$_{\!}$ annotations.

\begin{table}[t]
   \centering
   \resizebox{0.49\textwidth}{!}{
   \setlength\tabcolsep{2pt}
   \renewcommand\arraystretch{1.03}
   \begin{tabular}{|c||ccc|ccc|}
   \hline
   Method & mAP\! $\uparrow$& AP$_{50}$\! $\uparrow$ & AP$_{75}$\! $\uparrow$  & Precision\! $\uparrow$  & Recall\! $\uparrow$ & F-Score\! $\uparrow$ \\
   \hline
   \hline
     Seide \etal~\cite{seidel2019improved} &16.1& 39.4 &9.0 & 70.9 & 38.6& 50.0  \\
     Li \etal~\cite{li2019supervised} & 25.2& 69.9  & 30.2& 81.4 & 64.5 & 72.0 \\
     Tamura \etal~\cite{tamura2019omnidirectional} & 28.8& 59.8 &24.2 & 77.0  & 52.4 & 62.4 \\
     RAPiD~\cite{duan2020rapid} & 37.7& 72.0 & 26.8& 73.3 &67.8 & 70.4 \\\hline
     Ours & \textbf{45.4}  & \textbf{85.1} & \textbf{36.2}& \textbf{84.7}&\textbf{74.4} & \textbf{79.5} \\
    \hline
   \end{tabular}
   }
   \captionsetup{font=small}
   \caption{\small\textbf{Person detection results} on WEPDTOF~\cite{tezcan2022wepdtof} (\S\ref{sec:exPD}).}
   \label{table:fish_det2}
   \vspace{-12pt}
\end{table}

\begin{table*}[t]
   \centering
   \resizebox{0.99\textwidth}{!}{
   \setlength\tabcolsep{7pt}
   \renewcommand\arraystretch{1.03}
   \begin{tabular}{|c ||c|ccc|cc||c|ccc|cc|}
   \hline
   % \thickhline
   % \rowcolor{mygray}
   & \multicolumn{6}{c||}{\texttt{val}} & \multicolumn{6}{c|}{\texttt{test}} \\
   \cline{2-13}
   % \rowcolor{mygray}
   \multirow{-2}{*}{Method} & mPE\! $\downarrow$ &PE$_{n}$\! $\downarrow$& PE$_{m}$\! $\downarrow$ & PE$_{f}$\! $\downarrow$ &  PE$_{seen}$\! $\downarrow$ & PE$_{unseen}$\! $\downarrow$ & mPE\! $\downarrow$ &PE$_{n}$\! $\downarrow$& PE$_{m}$\! $\downarrow$ & PE$_{f}$\! $\downarrow$ &  PE$_{seen}$\! $\downarrow$ & PE$_{unseen}$\! $\downarrow$  \\
   \hline
   \hline
   Seide \etal~\cite{seidel2019improved} & 1.298 &  0.561 &1.332& 3.109 &1.206 &1.382
   &1.321 &0.706& 1.309& 3.482 &1.306 &1.386\\
    Li \etal~\cite{li2019supervised}& 0.898 & 0.502& 0.871& 2.650 & 0.832& 0.962
    &0.913& 0.543& 0.884 &2.780 & 0.904 &0.998  \\
    Tamura \etal~\cite{tamura2019omnidirectional} &0.755 & 0.429 & 0.736& 1.862& 0.709& 0.821&
    0.778 &0.471 &0.826 &2.160 & 0.724 & 0.836 \\
    RAPiD~\cite{duan2020rapid}  &  0.674 & 0.426 & 0.623 & 1.403& 0.625& 0.757
    & 0.682 & 0.461 &0.664 & 1.445& 0.638& 0.776 \\\hline
    Ours & \textbf{0.387}   & \textbf{0.164} & \textbf{0.382} & \textbf{0.786} & \textbf{0.332} & \textbf{0.419}  & \textbf{0.392} & \textbf{0.171} & \textbf{0.391}& \textbf{0.825}& \textbf{0.343} & \textbf{0.413} \\
   %\hline
   \hline
   \end{tabular}
   }
   \captionsetup{font=small}
   \caption{\small{\textbf{Person localization results} on \texttt{val} and \texttt{test} sets of LOAF (\S\ref{sec:exPL}). }}
    \label{table:loaf_loc}
   \vspace{-12pt}
\end{table*}

\noindent\textbf{Evaluation Metrics.} For fisheye person detection, we follow COCO~\cite{lin2014microsoft} to report the mean average precision (mAP) for IoU\!~$\in$\!~[0.5:~\!0.05:~\!0.95]. We also employ AP$_{50}$ and AP$_{75}$ for$_{\!}$ further$_{\!}$ analysis.$_{\!}$ 
For$_{\!}$ fisheye$_{\!}$ person$_{\!}$ localization,$_{\!}$ we mea- sure positional error (PE) in meters,  \ie, Euclidean distance of calculated position and ground-truth position, as the conventions in visual localization~\cite{li2012worldwide,sattler2017large,xiao2018indoor,zhu2019object}. For detailed evaluation, on our LOAF, we report performance \wrt horizontal distance between human and the fisheye lens, \ie, \textit{near} (0$\sim$10 m),  \textit{middle}$_{\!}$ (10$\sim$20 m),$_{\!}$ and$_{\!}$ \textit{far} (larger$_{\!}$ than$_{\!}$ 20 m). Hence we have AP$_{\{n,m,f\}}$ and PE$_{\{n,m,f\}}$ accordingly.

\subsection{Performance on Person Detection}\label{sec:exPD}
\noindent\textbf{LOAF.} We first report the person detection performance on LOAF.$_{\!}$ Specifically,$_{\!}$ four$_{\!}$ recent$_{\!}$ deep$_{\!}$ learning$_{\!}$ based$_{\!}$ fisheye person detectors~\cite{seidel2019improved,li2019supervised,tamura2019omnidirectional,duan2020rapid} are involved for compari- son.$_{\!}$ Among$_{\!}$ them,$_{\!}$ \cite{seidel2019improved,li2019supervised}$_{\!}$ are$_{\!}$ training-free,$_{\!}$ thus$_{\!}$ their$_{\!}$ scores are obtained by directly running the algorithms on the \texttt{val} and  \texttt{test} data. As \cite{tamura2019omnidirectional,duan2020rapid} are trainable methods, we first train them on the \texttt{train} set of LOAF following their offi- cial setups, and then report the scores on the \texttt{val} and  \texttt{test} sets. Quantitative comparison results are presented in Table~\ref{table:loaf_det}. Some essential observations are as follows:
\begin{itemize}[leftmargin=*]
   \setlength{\itemsep}{0pt}
   \setlength{\parsep}{-2pt}
   \setlength{\parskip}{-0pt}
   \setlength{\leftmargin}{-6pt}
   \vspace{-2pt}%In particular, semantic
   \item Our fisheye person detector  significantly outperforms existing methods across all the metrics. For example, our detector provides a considerable performance gain in mAP, \ie, \textbf{6.9\%} and \textbf{7.0\%} higher than the second best, RAPiD~\cite{duan2020rapid}, on the \texttt{val} and \texttt{test} sets, respectively.

   \item  Our detector not only handles nearby persons but also approaches distant targets well. Specifically, for the targets at 10$\sim$20 m away from the fisheye lens, our performance gain over other methods is more significant than the improvement reported with the persons at 0$\sim$10 m distance.

   \item Our detector also yields small performance gap between the \texttt{seen} and \texttt{unseen} scenes (\eg, 49.3 $\rightarrow$ 44.9 in terms of mAP), validating our good generalizability.

    \item For those persons at very far distance ($>$20 m), our algorithm, though still giving higher scores than other competitors, suffers from great performance degradation. This sheds light on the direction of our future efforts.
   \vspace{-4pt}
\end{itemize}

\noindent \textbf{CEPDOF$_{\!}$~\cite{duan2020rapid}.$_{\!}$} CEPDOF$_{\!}$ contains$_{\!}$ eight$_{\!}$ videos$_{\!}$ with$_{\!}$ 25,504 frames$_{\!}$ in$_{\!}$ total.$_{\!}$ Following$_{\!}$ the$_{\!}$ official$_{\!}$ setup,$_{\!}$ we$_{\!}$ train$_{\!}$ our fisheye$_{\!}$ person$_{\!}$ detector$_{\!}$ on$_{\!}$ HABBOF$_{\!}$~\cite{li2019supervised}$_{\!}$ and$_{\!}$ MW-R$_{\!}$~\cite{duan2020rapid} datasets,$_{\!}$ and$_{\!}$ report$_{\!}$ performance$_{\!}$ on$_{\!}$ CEPDOF.$_{\!}$ As$_{\!}$ shown$_{\!}$ in$_{\!}$ Table$_{\!}$~\ref{table:fish_det1}, our fisheye detector delivers state-of-the-art performance: it significantly outperforms RAPiD\!~\cite{duan2020rapid}, the current top-leading algorithm, by \textbf{7.5\%} in terms of mAP.

\noindent \textbf{WEPDTOF$_{\!}$~\cite{tezcan2022wepdtof}.}$_{\!}$ WEPDTOF has 16 videos with 10,544 frames in total. Following the official setup, we use HABBOF~\cite{li2019supervised}, MW-R~\cite{duan2020rapid}, and CEPDOF$_{\!}$~\cite{duan2020rapid} for training and WEPDTOF for testing. Table~\ref{table:fish_det2} summarizes the results. Impressively, our detector greatly suppresses all the other competitors across all the evaluation metrics.

\subsection{Performance on Person Localization}\label{sec:exPL}
Then we study the person localization performance on LOAF. None of previous fisheye person detectors~\cite{seidel2019improved,li2019supervised,tamura2019omnidirectional,duan2020rapid} are aware of the task of person localization. We therefore follow the common practice~\cite{zhu2019object,del2021robust} in the field of visual localization:  for~\cite{seidel2019improved,li2019supervised,tamura2019omnidirectional,duan2020rapid}, we project centers of their detected human boxes on the 3D world, based on the same strategy of ours (\textit{cf}.~\S\ref{sec:approach_stage2}), for physical position estimation.  As shown in Table~\ref{table:loaf_loc}, our system produces much small localization errors in comparison with~\cite{seidel2019improved,li2019supervised,tamura2019omnidirectional,duan2020rapid}, \ie, \textbf{0.392} m \textit{vs} 0.682 m\!~\cite{duan2020rapid} and 0.773 m of \cite{tamura2019omnidirectional} on the \texttt{test} set of LOAF. In addition, for all the methods, the localization performance is declined as target distance increases, but our system suffers from the smallest drop. Moreover, the results suggest there is still large room for improvement, thus we hope that our dataset could encourage continuous efforts in this challenging task.

\begin{table}[t]
\centering
\vspace{4pt}
        \resizebox{0.49\textwidth}{!}{
      \setlength\tabcolsep{1.5pt}
      \renewcommand\arraystretch{1.03}
\begin{tabular}{|c||c|c|cc|c|cc|}
\hline
~ & & \multicolumn{3}{c|}{\texttt{val} \texttt{seen}} & \multicolumn{3}{c|}{\texttt{val} \texttt{unseen}} \\
\cline{3-8}
% \noalign{\smallskip}
\multirow{-2}{*}{Method} & \multirow{-2}{*}{mAP\! $\uparrow$} & mAP\! $\uparrow$ &AP$_{50}$\! $\uparrow$ &AP$_{75}$\! $\uparrow$ &mAP\! $\uparrow$ &AP$_{50}$\! $\uparrow$ &AP$_{75}$\! $\uparrow$ \\
\hline
\hline
 Baseline~\cite{liu2022dab}  & 43.1 & 46.4 &81.2& 48.4  & 40.9 & 76.2 & 38.9 \\
+ Rotation Aug. & 45.6 & 49.2  & 82.8& 52.8 & 43.2 & 78.9 &43.9 \\
% + Rotation Equ. &  &  & &  &  &  & - \\
+ Rotation Equ. & \textbf{47.2} & \textbf{50.6} & \textbf{83.7} & \textbf{54.6} & \textbf{45.5} & \textbf{80.4} & \textbf{47.1} \\
\hline
\end{tabular} 
}
   \vspace*{-2pt}
\captionsetup{font=small}
   \caption{\small{\textbf{Diagnostic results} on \texttt{val} set of LOAF (\S\ref{sec:exabs}). }}
    \label{table:exAB_val}
\vspace*{-9pt}
\end{table}

\vspace{4pt}
\subsection{Diagnostic Study} \label{sec:exabs}
\vspace{3pt}
Table~\ref{table:exAB_val} studies the efficacy of our rotation equivariant training strategy~(\S\ref{sec:approach_stage1}), on the \texttt{val} set of LOAF. Our baseline model~\cite{liu2022dab} (\textit{row} \#1) gains 43.1\% mAP. After adopting rotational data$_{\!}$ augmentation$_{\!}$ (\textit{row} \#2), the$_{\!}$ performance$_{\!}$ boosts$_{\!}$ by \textbf{2.5\%}$_{\!}$ mAP. By contrast, our rotation equivariant training brings much larger improvements over the baseline, \eg, \textbf{4.1\%} mAP gain. It is remarkable that, training with standard rotational data augmentation technique can be viewed as a specific case of our equivariant training -- only learning rotation-equivariant object-querying (\textit{cf}.~Eq.~\ref{equ:rotate2}).

\vspace{5pt}
\section{Conclusion}
\vspace{2pt}
We presented LOAF -- the first top-view fisheye dataset that supports large-scale study for person localization in realistic surveillance scenarios.
With radius-aligned person-box annotations and precise location ground-truths, LOAF closes a crucial gap in the literature as these cases are not covered by previous datasets and annotation protocols. We further proposed an efficient fisheye person detection model that is equipped with a {rotation-equivariant} training strategy. The physical locations of detected persons are formulated based on the fisheye model and the altitude of the camera. We empirically verified the effectiveness and promising performance of our algorithm.

\clearpage
\appendix
\twocolumn[{
\maketitle
\begin{center}
\renewcommand\thetable{S1}
    \captionsetup{type=table}
     \vspace{-4pt}
   \resizebox{0.99\textwidth}{!}{
   \setlength\tabcolsep{4pt}
   \renewcommand\arraystretch{1.03}
   \begin{tabular}{|c|c||c|c|ccc|ccc|ccc|ccc|}
    \hline
   \multicolumn{2}{|c||}{}& \multirow{2}{*}{\# Video} & \multirow{2}{*}{\# Image}  & \multicolumn{3}{c|}{\# People} & \multicolumn{3}{c|}{Scene} & \multicolumn{3}{c|}{Season} & \multicolumn{3}{c|}{Time} \\\cline{5-16}
   \multicolumn{2}{|c||}{} & & & Total & Avg. & Max & Total & Indoor & Outdoor & Spring & Summer & Autumn & Morning & Noon & Afternoon \\
   \hline
   \hline
   \multicolumn{2}{|c||}{\texttt{train}} & 51 & 29,569 & 315,262 & 10.6 & 65 & 35 & 7 & 28 &13&30&8& 10 & 15 & 26 \\\hline 
    \multirow{3}{*}{\texttt{val}} & seen & 3 & 1,700 & 18,460 & 10.8 & 29 & 3 & 1 & 2 & 1&2 &0 & 1 & 1 & 1 \\
    & {\color{gray}unseen} & {\color{gray}5} & {\color{gray}2,900} & {\color{gray}29,381} & {\color{gray}10.1} & {\color{gray}44} & {\color{gray}5} & {\color{gray}2} & {\color{gray}3} & {\color{gray}1} & {\color{gray}3} & {\color{gray}1} & {\color{gray}1} & {\color{gray}2} & {\color{gray}2} \\
    & total& 8 & 4,600 & 47,841 & 10.4 & 44 & 8 & 3 & 5 & 2 & 5 & 1& 2 & 3 & 3 \\\hline
     \multirow{3}{*}{\texttt{test}} & seen & 5 & 2,774 & 28,666 & 10.3 & 41 & 5 & 2 & 3 & 1 & 3 & 1 & 1 & 2 & 2 \\
    & {\color{gray}unseen} & {\color{gray}10} & {\color{gray}5,999} & {\color{gray}65,993} & {\color{gray}10.0} & {\color{gray}44} & {\color{gray}10} & {\color{gray}2} & {\color{gray}8} & {\color{gray}3} & {\color{gray}5} & {\color{gray}2}& {\color{gray}3} & {\color{gray}3} & {\color{gray}4} \\
    & total& 15 & 8,773 & 94,659 & 10.1 & 44 & 15 & 4 & 11 & 4 & 8 & 3 & 4 & 5 & 6 \\\hline\hline
    \multicolumn{2}{|c||}{Total} & 74 & 42,942 & 457,762 & 10.5 & 65 & 50 & 11 & 39 & 17&38&11&14 & 20 & 32 \\\hline 
   \end{tabular}
   }
    \captionof{table}{Detailed statistics of LOAF. \# indicates the number of elements.}\label{table:attr}
\end{center}
}]

This document provides additional materials to supplement our main manuscript. We first present more statistics about LOAF in \S\ref{analysis}, and then give extra implementation details of our method in \S\ref{implem}. More qualitative results on the \texttt{test} set of LOAF are summarized  in \S\ref{quali}. Next, we state the ethical conducts in \S\ref{conducts}. Finally, we provide the pseudo of our proposed rotation equivariant training strategy in \S\ref{code}.

\section{Additional Dataset Analysis} \label{analysis} 
\noindent \textbf{More Statistics.}
LOAF is captured from multiple indoor/ outdoor scenes (\eg, library, classroom, street, parking lot) across three seasons, we summarize the detailed statistics in Table~\ref{table:attr}, including the number of boxes, video sequences, \etc. As seen, the majority of videos are collected from outdoor environments characterized by increased complexity, larger fields of view, and a higher number of human targets when compared to the indoor ones. These videos are divided into \texttt{train}, \texttt{val}, and \texttt{test} sets in the ratio of 7:1:2 respectively, while ensuring an roughly even distribution of attributes (\eg, season, time) across these sets.

\section{More Implementation Details} \label{implem}

\noindent \textbf{Training$_{\!}$ Objective.}$_{\!}$ We$_{\!}$ extend$_{\!}$ the$_{\!}$ Generalized$_{\!}$ IoU (GIoU) loss\!~\cite{Rezatofighi_2018_CVPR} utilized in vanilla DETR\!~\cite{carion2020end} for bounding box regression to the rotated setup. Concretely, Brute-force search is leveraged to compute the minimum enclosing box between two rotated bounding boxes. It is implemented in a fully differentiable manner and adapted for parallel processing on GPU, which merely defers the training speed by around 5\% when compared to the axis-aligned setup.

\section{Qualitative Evaluation} \label{quali}
\begin{figure*}  
\renewcommand\thefigure{S1} 
   \begin{center}
   \vspace{-5pt}
      \includegraphics[width=\linewidth]{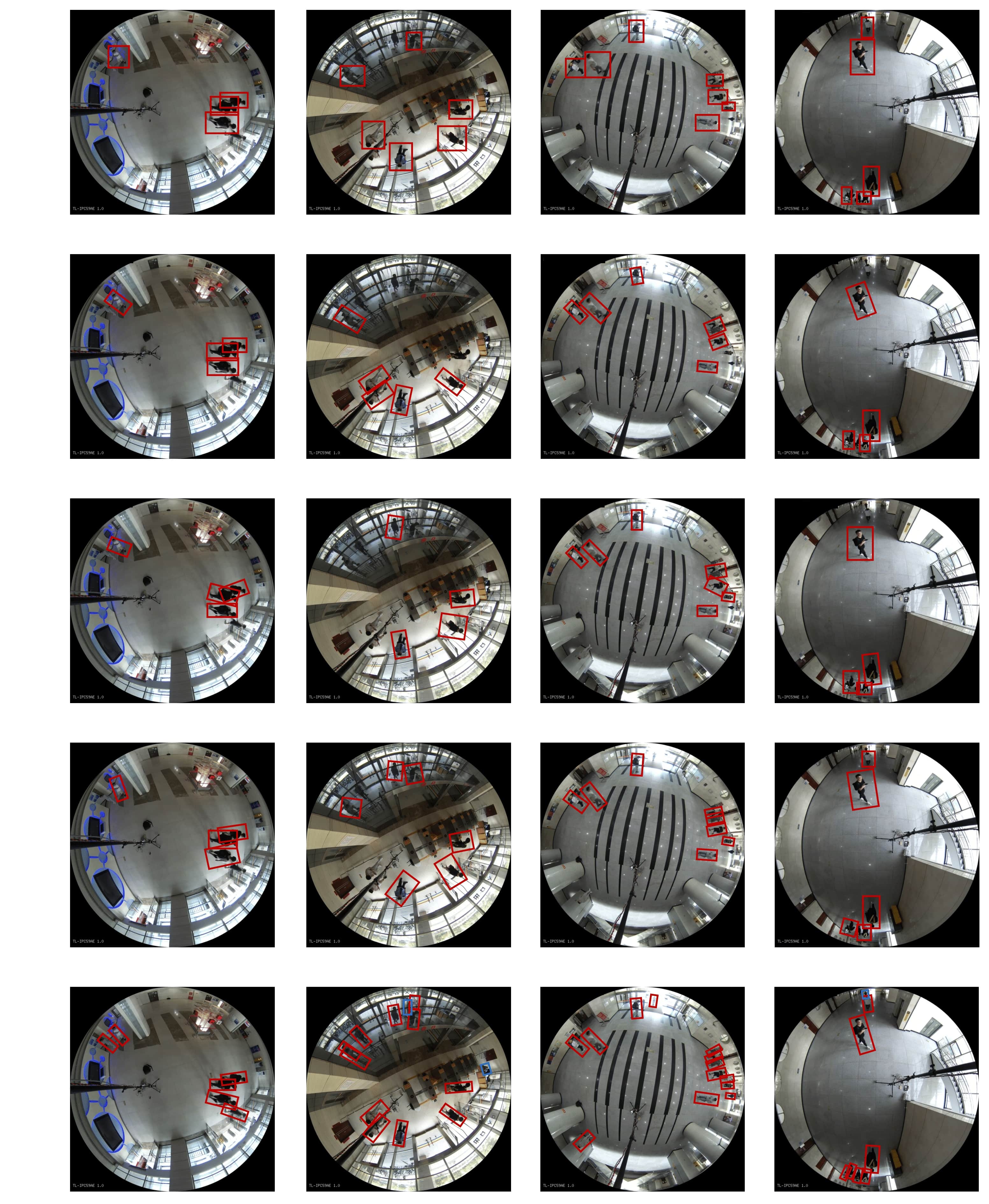}
      \put(-480.0,520){\rotatebox{90}{Seide \etal~\cite{seidel2019improved}}}
      \put(-480.0,406){\rotatebox{90}{Li \etal~\cite{li2019supervised}}}
      \put(-480.0,270){\rotatebox{90}{Tamura \etal~\cite{tamura2019omnidirectional}}}
      \put(-480.0,160){\rotatebox{90}{RAPiD~\cite{duan2020rapid}}}
      \put(-480.0,46){\rotatebox{90}{\textbf{Ours}}}
   \end{center}
   \vspace{-15pt}
   \captionsetup{font=small}
   \caption{\textbf{Visual comparison of detection results} on the \texttt{test} set of LOAF. \includegraphics[scale=0.075,valign=c]{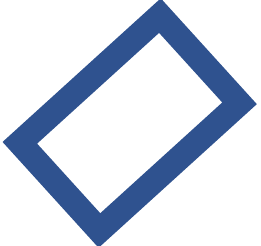} indicates targets missed by our method.}
   \label{fig:sm1}
   % \vspace{-15pt}
\end{figure*}

\begin{figure*}
\renewcommand\thefigure{S2}
   \begin{center}
   \vspace{-5pt}
      \includegraphics[width=\linewidth]{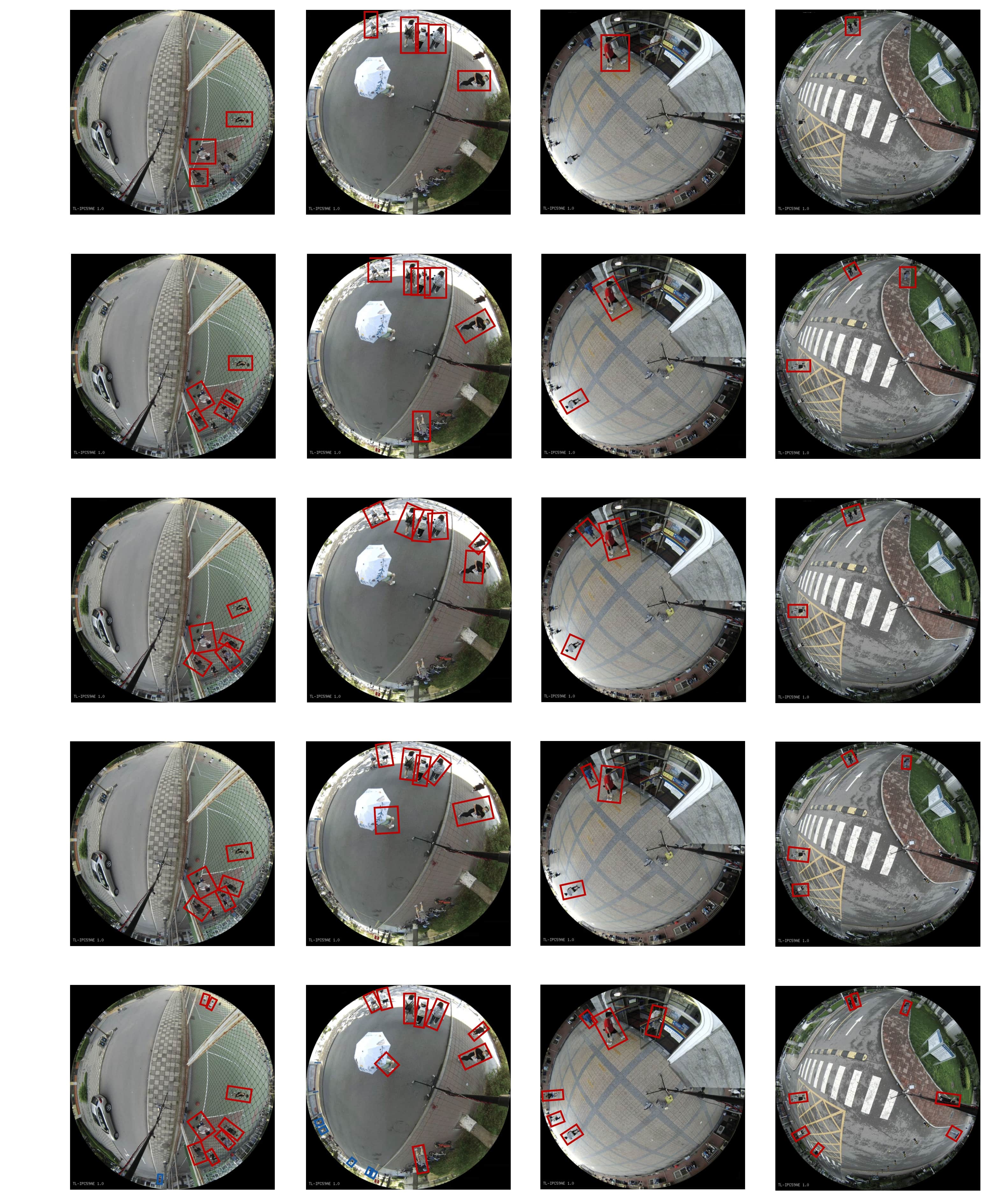}
      \put(-480.0,520){\rotatebox{90}{Seide \etal~\cite{seidel2019improved}}}
      \put(-480.0,406){\rotatebox{90}{Li \etal~\cite{li2019supervised}}}
      \put(-480.0,270){\rotatebox{90}{Tamura \etal~\cite{tamura2019omnidirectional}}}
      \put(-480.0,160){\rotatebox{90}{RAPiD~\cite{duan2020rapid}}}
      \put(-480.0,46){\rotatebox{90}{\textbf{Ours}}}
   \end{center}
   \vspace{-15pt}
   \captionsetup{font=small}
   \caption{\textbf{Visual comparison of detection results} on the \texttt{test} set of LOAF. \includegraphics[scale=0.075,valign=c]{figure/box} indicates targets missed by our method.}
   \label{fig:sm2}
   % \vspace{-15pt}
\end{figure*}

\begin{figure*}
\renewcommand\thefigure{S3}
   \begin{center}
   \vspace{-5pt}
      \includegraphics[width=\linewidth]{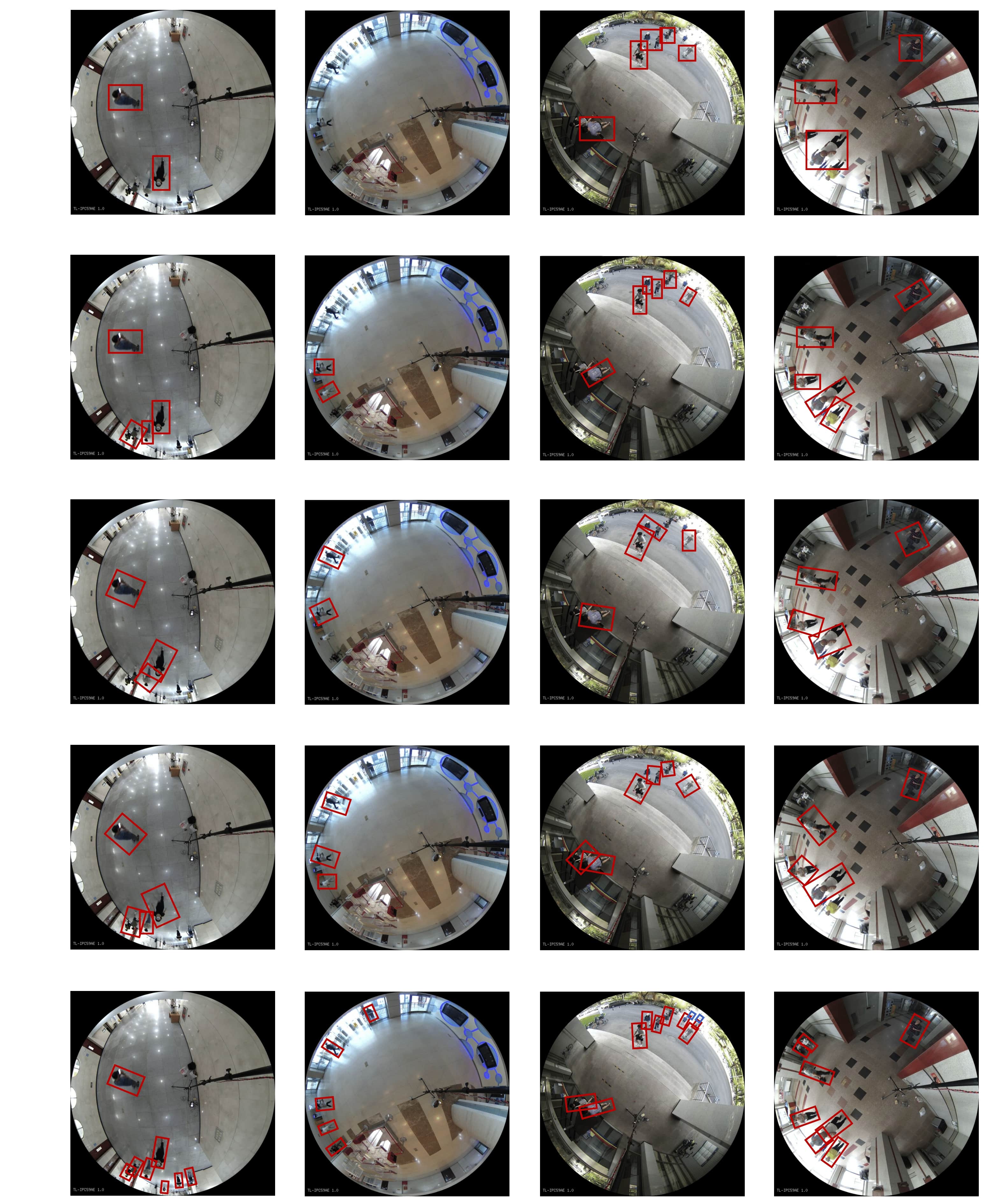}
      \put(-480.0,520){\rotatebox{90}{Seide \etal~\cite{seidel2019improved}}}
      \put(-480.0,406){\rotatebox{90}{Li \etal~\cite{li2019supervised}}}
      \put(-480.0,270){\rotatebox{90}{Tamura \etal~\cite{tamura2019omnidirectional}}}
      \put(-480.0,160){\rotatebox{90}{RAPiD~\cite{duan2020rapid}}}
      \put(-480.0,46){\rotatebox{90}{\textbf{Ours}}}
   \end{center}
   \vspace{-15pt}
   \captionsetup{font=small}
   \caption{\textbf{Visual comparison of detection results} on the \texttt{test} set of LOAF. \includegraphics[scale=0.075,valign=c]{figure/box} indicates targets missed by our method.}
   \label{fig:sm3}
   % \vspace{-15pt}
\end{figure*}

\begin{figure*}
\renewcommand\thefigure{S4}
   \begin{center}
   \vspace{-5pt}
      \includegraphics[width=\linewidth]{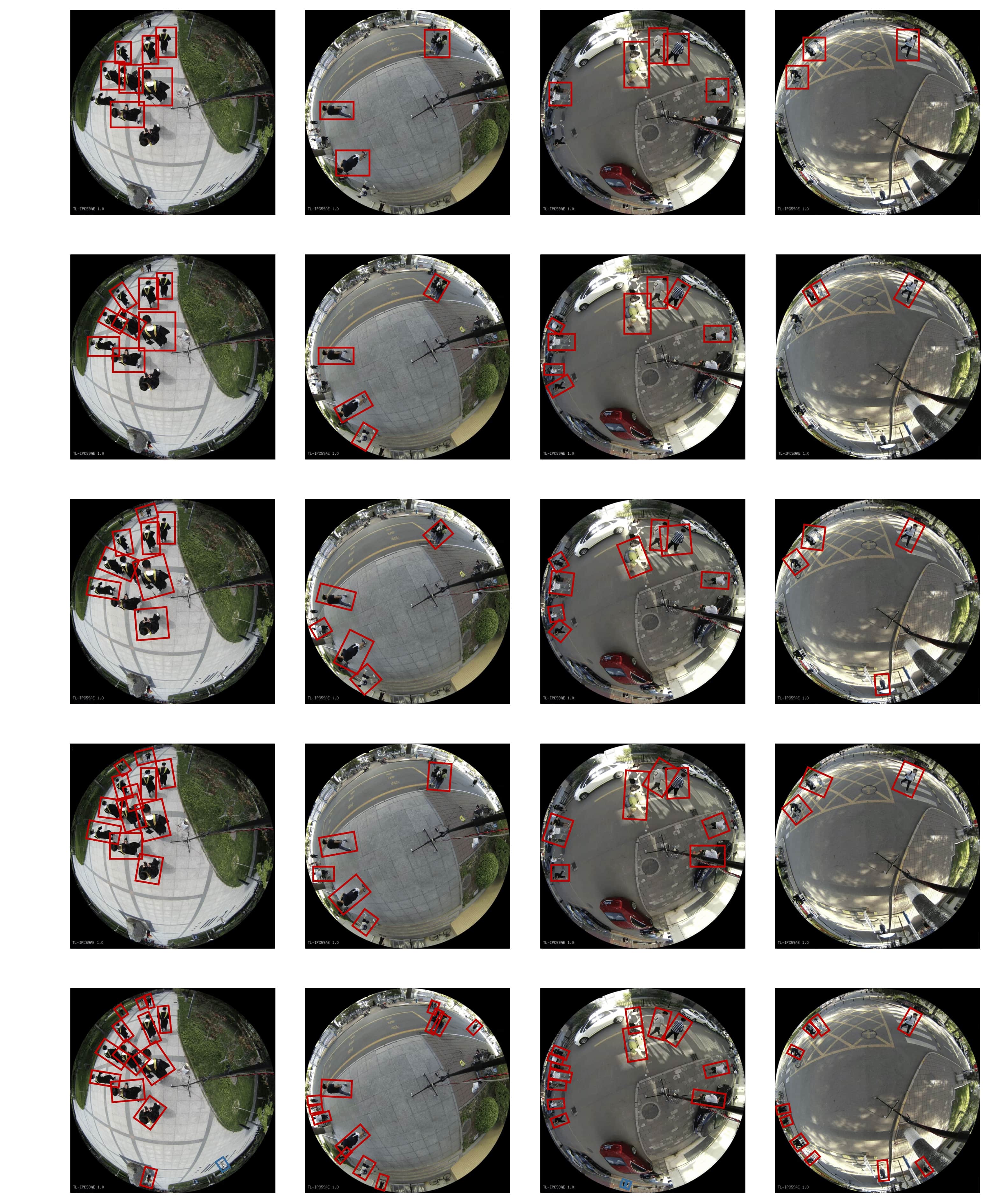}
      \put(-480.0,520){\rotatebox{90}{Seide \etal~\cite{seidel2019improved}}}
      \put(-480.0,406){\rotatebox{90}{Li \etal~\cite{li2019supervised}}}
      \put(-480.0,270){\rotatebox{90}{Tamura \etal~\cite{tamura2019omnidirectional}}}
      \put(-480.0,160){\rotatebox{90}{RAPiD~\cite{duan2020rapid}}}
      \put(-480.0,46){\rotatebox{90}{\textbf{Ours}}}
   \end{center}
   \vspace{-15pt}
   \captionsetup{font=small}
   \caption{\textbf{Visual comparison of detection results} on the \texttt{test} set of LOAF. \includegraphics[scale=0.075,valign=c]{figure/box} indicates targets missed by our method.}
   \label{fig:sm4}
   % \vspace{-15pt}
\end{figure*}

\begin{figure*}
\renewcommand\thefigure{S5}
   \begin{center}
   \vspace{-5pt}
      \includegraphics[width=0.88\linewidth]{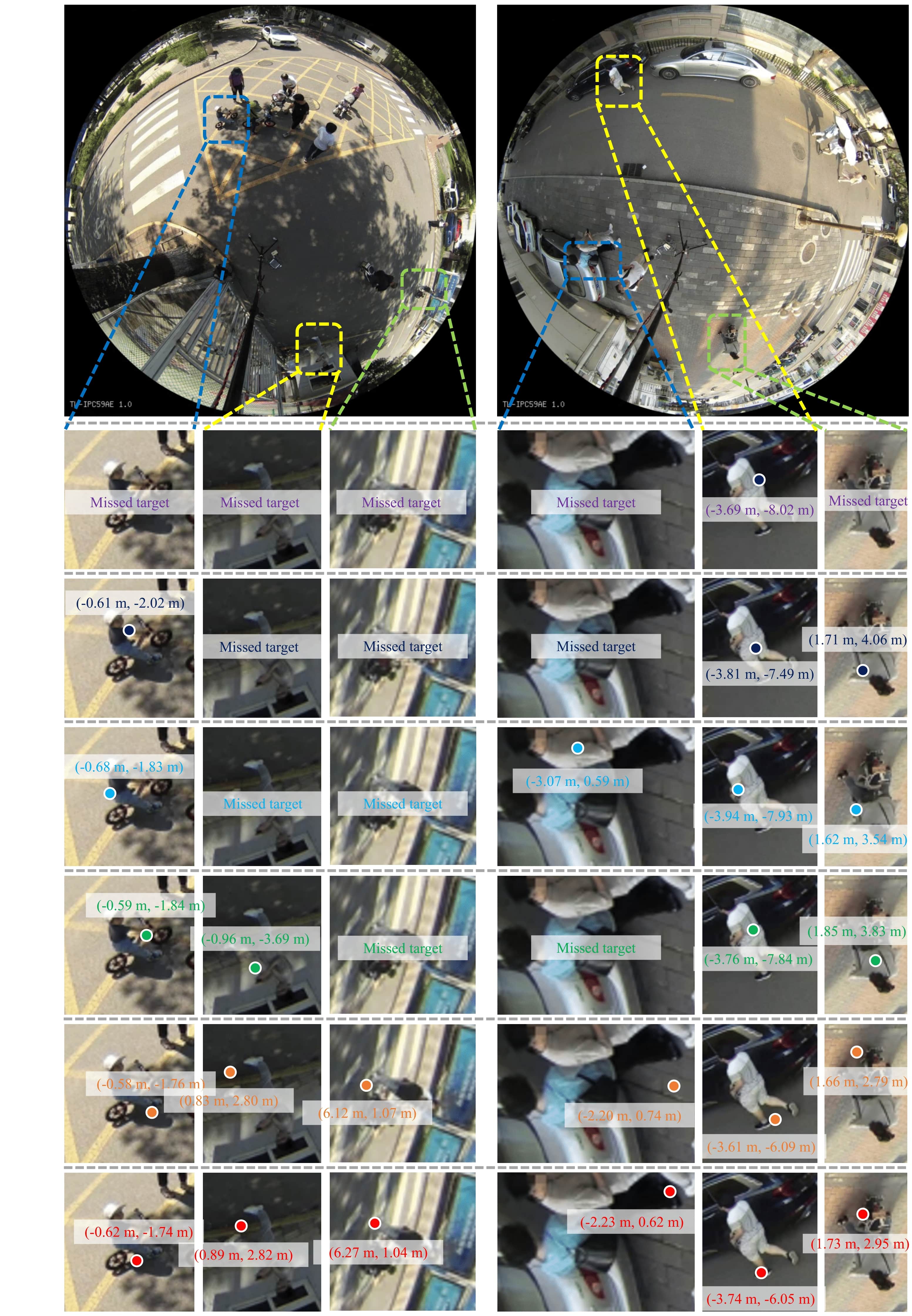}
      \put(-420.0,361){\rotatebox{90}{Seide \etal~\cite{seidel2019improved}}}
      \put(-420.0,297){\rotatebox{90}{Li \etal~\cite{li2019supervised}}}
      \put(-420.0,215){\rotatebox{90}{Tamura \etal~\cite{tamura2019omnidirectional}}}
      \put(-420.0,156){\rotatebox{90}{RAPiD~\cite{duan2020rapid}}}
      \put(-420.0,95){\rotatebox{90}{\textbf{Ours}}}
       \put(-420.0,9){\rotatebox{90}{Ground truth}}
   \end{center}
   \vspace{-15pt}
   \captionsetup{font=small}
   \caption{\textbf{Visual comparison of localization results} on the \texttt{test} set of LOAF. We selected three targets per frame for clear visualization.}
   \label{fig:sm5}
   % \vspace{-15pt}
\end{figure*}

\noindent \textbf{Visual Comparison.} Fig.\!~\ref{fig:sm1}-\ref{fig:sm5} compare our method with existing work qualitatively. It is obvious that our proposed method consistently presents more accurate detection and localization results, regardless of the category of scenes. Notably, it is much more effective than existing work for targets that are relatively small or densely arranged.

\begin{figure*}
\renewcommand\thefigure{S6}
   \begin{center}
   \vspace{-5pt}
      \includegraphics[width=0.99\linewidth]{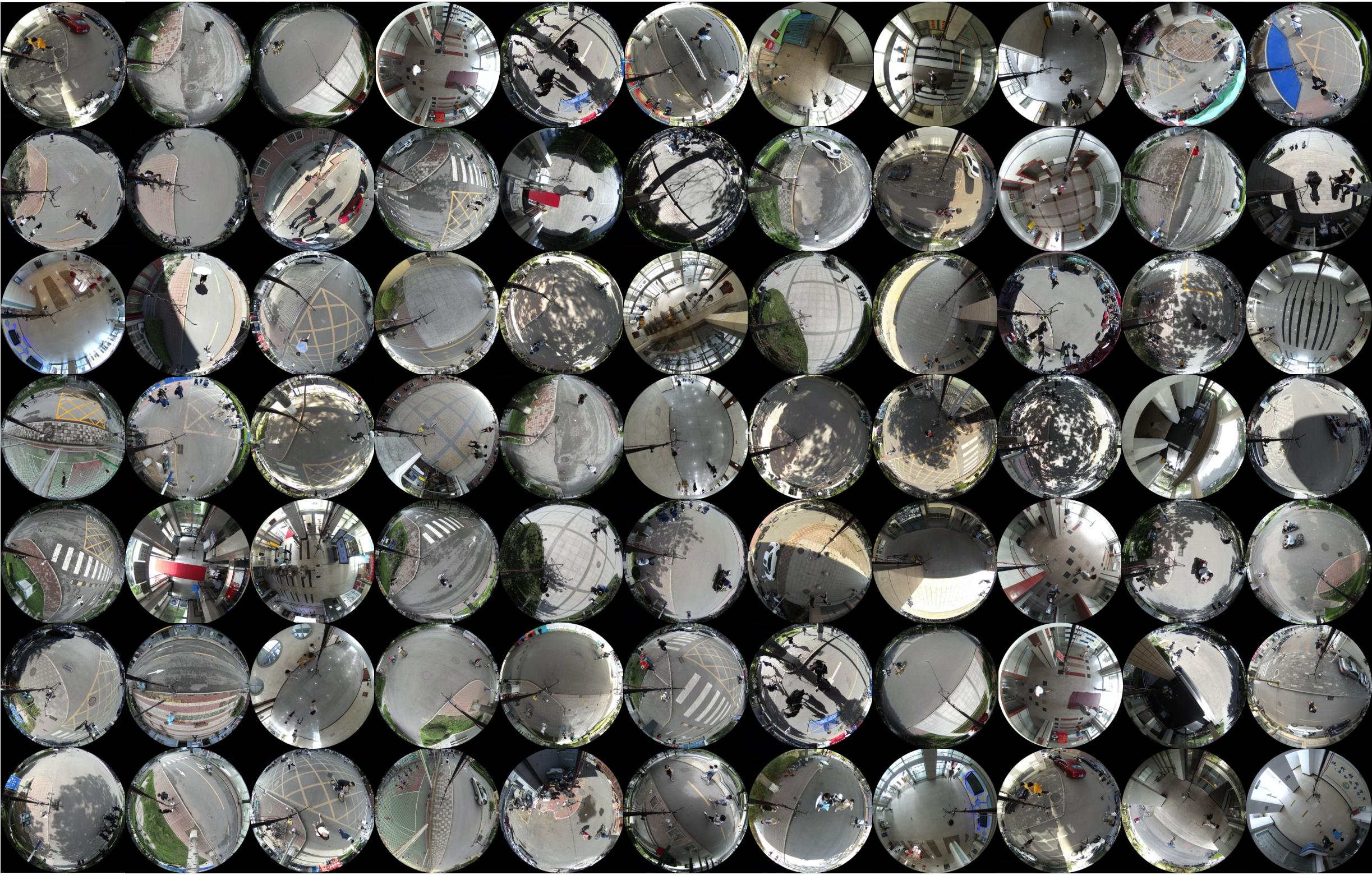}
   \end{center}
   \vspace{-15pt}
   \captionsetup{font=small}
   \caption{A collage constituted from various scenes characterized by distinct attributes.}
   \label{fig:sm6}
   % \vspace{-15pt}
\end{figure*}

\noindent \textbf{Diversity.} To render a more intuitive understanding of the diversity of LOAF, a collage constituted from various scenes characterized by distinct attributes is given in Fig.\!~\ref{fig:sm6}.

\section{Ethical Conducts} \label{conducts}
To protect the privacy of individuals and groups, we utilize Gaussian filters to blur all visible facial regions in LOAF. 
The proprietary data can only be accessed for non-commercial purposes to prevent inappropriate usage.

\section{Pseudo Code} \label{code}
We offer the pseudo code for our proposed query-based rotation equivariant training strategy in Algorithm \ref{alg:equ}. 

\begin{algorithm}
\renewcommand\thealgorithm{S1}
% \vspace{-10pt}
\caption{Pseudo-code for our proposed rotation equivariant training strategy.}
\label{alg:equ}
% \vspace{-10pt}
\definecolor{codeblue}{rgb}{0.25,0.5,0.5}
\lstset{
  backgroundcolor=\color{white},
  basicstyle=\fontsize{7.2pt}{7.2pt}\ttfamily\selectfont,
  columns=fullflexible,
  breaklines=true,
  captionpos=b,
  escapeinside={(:}{:)},
  commentstyle=\fontsize{7.2pt}{7.2pt}\color{codeblue},
  keywordstyle=\fontsize{7.2pt}{7.2pt},
%  frame=tb,
}
\begin{lstlisting}[language=python]
"""
I: input image 
gt: ground truth
angle: degree of clockwise rotation
(:\color{codegreen}{$\lambda$}:): the balance factor
"""
(:\color{codedefine}{\textbf{def}}:) (:\color{codefunc}{\textbf{rotat\_equi\_training}}:)(I, gt):
    # (:\color{codegreen}{$\mathcal{F}(I)$}:)
    m1 = (:\color{codepro}{\textbf{Encoder}}:)(I)
    angle = (:\color{codedim}{\textbf{randint}}:)(0, 360)
    # (:\color{codegreen}{$\mathcal{F}(g^r(I))$}:)
    m2 = (:\color{codepro}{\textbf{Encoder}}:)((:\color{codedim}{\textbf{rotate}}:)(I, angle))
   
    # (:\color{codegreen}{$\{\bm{q}_n\}_{n=1\!}^N\!=\!\kappa(\bm{I})$}:)
    query1 = (:\color{codedim}{\textbf{gen\_proposal}}:)(m1)
    # (:\color{codegreen}{$\{\bm{q}_n^{g^r\!}\}_{n=1\!}^N\!=\!\!\kappa(\bm{I}^{g^r\!})$}:)
    query2 = (:\color{codedim}{\textbf{gen\_proposal}}:)(m2)

    # (:\color{codegreen}{$\mathcal{D}(\bm{I}, \{\bm{q}_n\}_{n=1}^N)$}:)
    det1 = (:\color{codepro}{\textbf{Decoder}}:)(m1, query1)
    # (:\color{codegreen}{$\mathcal{D}(g^{r\!}(\bm{I}), \{\bm{q}_n^{g^r\!}\}_{n=1}^N)$}:)
    det2 = (:\color{codepro}{\textbf{Decoder}}:)((:\color{codedim}{\textbf{rotate}}:)(m1, angle), query2)

    # (:\color{codegreen}{$\mathcal{L}_\text{det}\big(\{\bm{b}_{\ell(n)}\}_{n=1}^{{N}}, \mathcal{D}(\bm{I}, \{\bm{q}_n\}_{n=1}^N)\big)$}:)
    loss1 = (:\color{codedim}{\textbf{det\_loss}}:)(det1, gt)
    # (:\color{codegreen}{$\mathcal{L}_\text{det}\big(\{\bm{b}^{g^r}_{\ell(n)}\}_{n=1}^{{N}}, \mathcal{D}(g^{r\!}(\bm{I}), \{\bm{q}_n^{g^r\!}\}_{n=1}^N)\big)$}:)
    loss2 = (:\color{codedim}{\textbf{det\_loss}}:)(det2, (:\color{codedim}{\textbf{label\_rotate}}:)(gt, angle)

    (:\color{codedefine}{\textbf{return}}:) loss1 + (:$\lambda$:)*loss2

\end{lstlisting} 
\end{algorithm}

\newpage
{\small
\bibliographystyle{ieee_fullname}
\bibliography{egbib}
} 

\end{document}